\documentclass[twoside]{article} 
\usepackage[accepted]{aistats2025}

\usepackage{enumitem}
\setlist[itemize]{noitemsep} 

\usepackage[round, authoryear]{natbib}
\renewcommand{\refname}{References} 

\usepackage{titlesec}
\titleformat{\section}
  {\normalfont\Large\bfseries}
  {\thesection}
  {1em}
  {\MakeUppercase}

\usepackage{amsmath}
\usepackage{amsthm}
\usepackage{thmtools, thm-restate}
\usepackage[utf8]{inputenc} 
\usepackage[T1]{fontenc}    
\usepackage{hyperref}       
\usepackage{url}            
\usepackage{booktabs}       
\usepackage{amsfonts}       
\usepackage{nicefrac}       
\usepackage{microtype}      
\usepackage{xcolor}         
\usepackage{graphicx}
\usepackage{wrapfig}

\usepackage[american]{babel}
\usepackage{amssymb}
\usepackage{cases}
\usepackage{standalone}
\standaloneconfig{mode=buildnew}
\usepackage{cleveref}

\theoremstyle{definition}
\newtheorem{definition}{Definition}[section]

\newtheorem{remark}{Remark}[section]

\newtheorem{conj}{Conjecture}[section]

\Crefname{prop}{Proposition}{proposition} 
\usepackage{verbatim}
\usepackage{algorithm}
\usepackage{algpseudocode}
\usepackage{adjustbox}
\PassOptionsToPackage{export}{adjustbox}
\usepackage{relsize}

\usepackage{float}
\usepackage{blindtext}
\usepackage{caption}
\usepackage{multirow}
\usepackage{fixltx2e,graphicx,mathpazo}
\usepackage{subfigure}
\usepackage{hyperref}
\usepackage{balance}

\usepackage{tikz} 

\usepackage{graphicx}

\usepackage{thumbpdf,lmodern}
\usepackage{placeins}
\usepackage[colorinlistoftodos]{todonotes}
\usepackage{framed}

\newcommand{\sreachable}{s\mathcal{R}}
\newcommand{\vreachable}{v\mathcal{R}}

\newcommand{\mnondom}[1]{\ensuremath{\mathcal{C}_{\ell(\cdot), R(\cdot)}\left(#1, f_{\theta}(\cdot),\theta^{(0)}\right)}}

\newcommand{\PerfInd}{{O}}
\newcommand{\hyperv}{e\mathcal{HV}}

\newcommand{\mnondomfront}[1]{\ensuremath{\mathcal{E}_{\ell(\cdot), R(\cdot)}\left(#1,f_{\theta}(\cdot),\theta^{(0)}\right)}}

\newcommand{\printfnsymbol}[1]{%
\textsuperscript{\@fnsymbol{#1}}%
}

\newcommand{\setpoint}{b}

\newcommand{\mytitle}{M-HOF-Opt: Multi-Objective Hierarchical Output Feedback Optimization via Multiplier Induced Loss Landscape Scheduling 
}

\begin{document}
\runningtitle{M-HOF-Opt: Multi-Objective Hierarchical Output Feedback Optimization}
\runningauthor{Xudong Sun, et~al.}

\twocolumn[
\aistatstitle{\mytitle}
\aistatsauthor{
  Xudong Sun$^{1}$ \And
  Nutan Chen$^{2}$ \And
  Alexej Gossmann$^{5}$ \And
  Matteo Wohlrapp$^{4}$
}
\aistatsauthor{
  Yu Xing$^{3}$ \And
  Emilio Dorigatti$^{1}$ \And
  Carla Feistner$^{6}$ \And
  Felix Drost$^{1}$
}
\aistatsauthor{
  Daniele Scarcella$^{1}$ \And
  Lisa Helen Beer$^{4}$ \And
  Carsten Marr$^{1}$
}

\aistatsaddress{
  $^{1}$Institute of AI for Health, Computational Health Center, Helmholtz Munich, Germany 
  \\
  $^{2}$Machine Learning Research Lab, Volkswagen Group, Munich, Germany
  \\
  $^{3}$Decision and Control, EECS, KTH Royal Institute of Technology, Stockholm, Sweden
  \\
  $^{4}$Technical University Munich, Germany
  \\
  $^{5}$U.S. FDA/CDRH, Silver Spring, MD, USA (work conducted here, no longer affiliated)
  \\
  $^{6}$ Applied Geology, GeoZentrum Nordbayern, Friedrich-Alexander-Universität, Erlangen-Nürnberg, Germany
}
] 

\begin{abstract}
A probabilistic graphical model is proposed, modeling the joint model parameter and multiplier evolution, with a hypervolume based likelihood, promoting multi-objective descent in structural risk minimization. 
We address multi-objective model parameter optimization via a surrogate single objective penalty loss with time-varying multipliers, equivalent to online scheduling of loss landscape. 
The multi-objective descent goal is dispatched hierarchically into a series of constraint optimization sub-problems with shrinking bounds according to Pareto dominance. The bound serves as setpoint for the low-level multiplier controller to schedule loss landscapes via output feedback of each loss term. Our method forms closed loop of model parameter dynamic, circumvents excessive memory requirements and extra computational burden of existing multi-objective deep learning methods, and is robust against controller hyperparameter variation, demonstrated on domain generalization tasks with multi-dimensional regularization losses.

\end{abstract}

\section{Introduction}
This work initializes efforts to utilize feedback mechanism, constraint optimization, hierarchical and optimal control into deep learning by treating neural network training as a dynamic process. We tackle two fold challenges in multi-objective deep learning and automatic multiple-multiplier adaptation in structural risk minimization: In many deep learning fields including domain generalization~\citep{gulrajani2020search}, the loss function for neural networks amounts to a summation of multiple terms, with a multiplier weighting each term. As the number of loss terms increases, the space of choice for those multipliers grows dramatically. In addition, when the validation set is drawn from a different distribution than the target domain, the validation performance does not faithfully reflect the target domain performance~\citep{chen2022pareto} and can easily reach and maintain a saturated value during training. All these factors complicate and add difficulties to the process of model selection using conventional hyperparameter tuning \citep{probst2019tunability}.
Inspired by feedback control theory~\citep{doyle2013feedback} and feedback optimization in power systems~\citep{picallo2020closing,he2023model,hauswirth2024optimization}, we treat the dynamic system of model parameters driven by the low level optimization algorithm as an uncontrolled plant, with the model parameters as state and the many loss terms as output. We use
state-dependent multiplier induced loss landscapes (see~\Cref{fig:landscape_scheduling}) as control input and designed 
a hierarchical control structure~\citep{dietterich1998maxq,kulkarni2016hierarchical,sun2020reinbo} to drive the system towards a multi-objective descent goal, without modifying the internal dynamic of low level neural network optimization algorithms.


We demonstrate that our method removes the combination curse of dimensionality of multi-dimensional multipliers. Our method only has a small number of hyperparameters for our controllers while achieving robust performance when changing them. 
We attribute the improved performance of our method to the automatic trade-off of different loss terms. 

Our major contributions are:
\begin{itemize}
\item We propose a novel multi-objective optimization algorithm via constraint optimization with shrinking bound. See~\Cref{prop:pareto_descent_constrained}. 
\item  We propose a probabilistic graphical model for depicting the joint model parameter and multiplier adaptation process with a hypervolume based likelihood, promoting multi-objective descent of each loss term, as an alternative model selection criteria in absence of trustworthy validation data.
\item We bring hierarchical control to solve the joint model parameter and multiplier inference of the probabilistic graphical model as a sequential decision process through an optimal control formulation and optimize the multi-objective descent goal hierarchically by breaking the goal into a series of constraint optimization sub-goals.
  \item We bring feedback control theory into deep learning via treating the neural network model parameter dynamic system driven by low level optimization algorithm as an uncontrolled plant. We propose to use state-dependent multiplier-induced loss landscapes as control input and offer initial theoretical analysis of the multi-objective descent behavior of our closed loop system.
\end{itemize}

\section{Preliminaries}\label{sec:preliminary}
\subsection{Structural risk minimization}\label{sec:prerequisite_srm}
Structural Risk Minimization (SRM) is a principle in learning algorithms that balances training performance and model complexity via optimizing the penalized loss in \Cref{eq:srm}.
\begin{align}
\begin{split}
L(\theta, \mu, \mathcal{D}_{tr}) &=\ell(\theta, \mathcal{D}_{tr}) + \mu^T R(\theta, \mathcal{D}_{tr})
\label{eq:srm}\\
\mu, R(\cdot) &\in \mathbb{R}_{+}^d 
\end{split}
\end{align}
In \Cref{eq:srm}, we consider loss $L$ with training data $\mathcal{D}_{tr}$ and model parameters $\theta$. $\ell(\theta, \mathcal{D}_{tr})$ represents the empirical risk and $R(\theta, \mathcal{D}_{tr})$ represents the regularization term of the loss function, also referred to as the penalty function, and thus $L$ the penalized loss. $\mathbb{R}_{+}^d$ is the $d$ dimensional positive octant. 
Here $\mu$ is the penalty (weight) multiplier, a special case of general hyperparameters (e.g.~learning rate) in deep learning.
\subsection{Notation}\label{sec:notation}
In the following sections, for brevity, we use $\ell(\cdot)$ and $R(\cdot)$ to indicate the implicit dependence of $\ell$ and $R$ on $\mathcal{D}_{tr}$ and $\theta$. We also write $R(\theta, \cdot)$ where we need an explicit dependence on $\theta$ and use $\cdot$ to implicitly represent other arguments like $\mathcal{D}_{tr}$. 
We use subscript to indicate the component of $R(\cdot)$, e.g., if $\mu=[\beta, \gamma]$, then $R_{\beta}(\cdot)$ corresponds to the component of $R(\cdot)$ weighted by $\beta$.
We use superscript $k$ in bracket to index the optimization iteration (See~\Cref{remark:pgm_bo_realization,remark:pgm_bo_fbopt_realization}). 

\begin{definition}[Model parameter dynamic system]\label{def:model_para_dyn}
When optimizing \Cref{eq:srm} with multiplier $\mu$ iteratively, 
we use $\theta^{(k+1)}=f_{\theta}(\mu, \theta^{(k)}, \mathcal{D}_{tr}, \ell(\cdot), R(\cdot))$ to represent the map bringing $\theta^{(k)}$ to its next value $\theta^{(k+1)}$. 
\end{definition}
Following $R(\cdot), \ell(\cdot)$, 
we have $f_{\theta}(\theta, \cdot)$, $f_{\theta}(\mu, \cdot)$ and $f_{\theta}(\cdot)$ to implicitly represent omitted arguments of $f_{\theta}$. We use $\theta^{+}$ for new value of $\theta$ after an operation without explicitly stating how many iterations are needed, resulting in notation $\theta^{+}=f_{\theta}(\mu, \theta, \cdot)$ and $\theta^{+}=f_{\theta}(\theta, \cdot)$.
\subsection{Multi-objective optimization}\label{sec:pre_multiobj_opt}
\begin{definition}[$R$ dominance and non-dominance]\label{def:non_dominance}
We use $R(\theta_1, \cdot) \prec R(\theta_2, \cdot)$ (see \Cref{eq:srm}) to indicate each component of the $d$ dimensional vector $R(\theta_1, \cdot)$ is $\le$ the corresponding component of $R(\theta_2, \cdot)$. We use $\nprec$ as the negation of $\prec$. The clause $\ell(\theta_1, \cdot), R(\theta_1, \cdot) \nprec \ell(\theta_2, \cdot), R(\theta_2, \cdot)$ AND $\ell(\theta_2, \cdot),R(\theta_2, \cdot) \nprec \ell(\theta_1, \cdot),R(\theta_1, \cdot)$ defines an equivalence relation which we denote as $\theta_1 \sim\equiv \theta_2$.
\end{definition}
\begin{definition}[Reachability set and value]\label{def:reachable_set}
Starting from $\theta^{(0)}$, under dynamic system $\theta^{+}=f_{\theta}(\theta, \cdot)$, we use $\sreachable\left(\theta^{(0)}, f_{\theta}(\cdot)\right)$ to represent the set of model parameters ($\theta$ points) that can be reached. Accordingly, we use $\vreachable_{\ell(\cdot), R(\cdot)}\left(\theta^{(0)}, f_{\theta}(\cdot)\right)$ to represent the corresponding set of multi-objective function values. 
\end{definition}
\begin{definition}[Non-dominant set map]\label{def:pareto_ec}
If $\theta_1\sim\equiv\theta_2$, where $\theta_1\in\sreachable(\theta^{(0)}, f_{\theta}(\cdot))$ and $\theta_2\in\sreachable(\theta^{(0)}, f_{\theta}(\cdot))$ as in \Cref{def:reachable_set},
we define $\theta_2\in \mnondom{\theta_1}$ the map from a representing element $\theta_1$ to its equivalence class, which contains $\theta_2$. We use $\mnondomfront{\theta_1}$
to represent the set of $\ell(\cdot), R(\cdot)$ values 
for each element of $\mnondom{\theta_1}$
. 
\end{definition}

\subsection{Conventional multiplier tuning}\label{sec:hyperopt}
The multiplier $\mu$ in \Cref{eq:srm} is a hyperparameter that affects the evolution of the model parameters $\theta$ in the optimization process for $L$ defined in \Cref{eq:srm}, which leads to optimized model parameter $\theta_{\mu, \mathcal{D}_{tr}}$ in~\Cref{eq:ml_theta}, where we omit the potential dependence of $\theta_{\mu, \mathcal{D}_{tr}}$ on its initial condition $\theta^{(0)}$ for notational simplicity.
\begin{align}
\theta_{\mu, \mathcal{D}_{tr}}
&=\arg\min_{\theta}L(\theta, \mu, \mathcal{D}_{tr}).\label{eq:ml_theta}
\end{align} 
Hyperparameter optimization aims to adjust the multiplier $\mu$ in alignment with specific performance metrics $\PerfInd$ 
(e.g.~validation set accuracy for a classification task) in \Cref{eq:hyperopt} evaluated on the validation set $\mathcal{D}_{val}$, which leads to the selected hyperparameter $\mu_{\mathcal{D}_{tr},\mathcal{D}_{val}}$.
\begin{align}
\mu_{\mathcal{D}_{tr},\mathcal{D}_{val}} =\arg\min_{\mu} O(\theta_{\mu, \mathcal{D}_{tr}}, \mathcal{D}_{val}).\label{eq:hyperopt}
\end{align}
The process of hyperparameter optimization in \Cref{eq:hyperopt} leads to selected model $\theta_{\mu_{\mathcal{D}_{tr},\mathcal{D}_{val}},\mathcal{D}_{tr}}$, which can be conceptualized as a problem of algorithm configuration~\citep{hutter2007automatic,hutter2014aclib, lopez2016irace, sun2020reinbo} where the hyperparameter configures a machine learning algorithm.

To solve \Cref{eq:hyperopt} iteratively, each iteration relies on a complete training cycle to optimize \Cref{eq:ml_theta}. The selection of $\mu$ depends on validation set $\mathcal{D}_{val}$, which can only be drawn from a different distribution compared to the target domain for the domain generalization problem (\Cref{sec:preliminary_dg}). Additionally, the performance metric $O$ on the in-domain validation set can reach and maintain a saturated value during training with neural networks of high expressive power. 


\section{Methods}\label{sec:method}
\begin{figure}[!ht]
\centering
\includegraphics[width=0.6\linewidth]{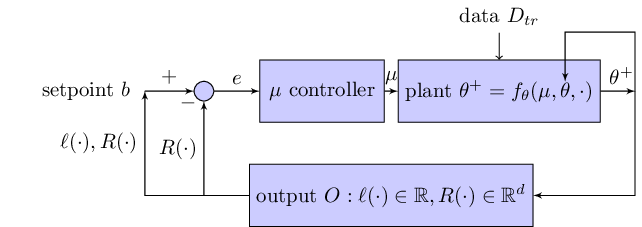}\caption{Control diagram illustrating the hierarchical output feedback optimization process with multi-objective setpoint adaptation. The uncontrolled plant corresponds to an open loop dynamic $\theta^{+}=f_{\theta}(\mu, \theta, \cdot)$ defined in~\Cref{def:model_para_dyn}, via optimizing \Cref{eq:srm} with a low level optimization algorithm lacking feedback. The $\mu$ controller adjusts the multiplier $\mu$, thus schedules the loss landscape as illustrated in~\Cref{fig:landscape_scheduling}, based on the difference $e$ between the setpoint $b$ and the measured output components $R(\cdot)$, guiding the optimization of model parameters $\theta$ through a feedback loop. 
The setpoint $b$ is adjusted via feedback from $\ell(\cdot)$ and $R(\cdot)$,
which forms a higher hierarchy. See~\Cref{fig:hyper_para_tune_bayes} for the probabilistic description of the closed loop behavior of this control diagram.
}\label{fig:control_diagram_fbopt}
\end{figure}

In this section, we elaborate on our multi-objective hierarchical output feedback optimization (M-HOF-Opt) as a control strategy in~\Cref{fig:control_diagram_fbopt} for the multidimensional multiplier adaptation in \Cref{eq:srm}. We propose a probabilistic graphical model for the joint model parameter and multiplier adaptation process in \Cref{sec:pgm}, which leads to an optimal control formulation of parameter estimation, extending the classical hyperparameter tuning-based approach to parameter-multiplier co-evolution in \Cref{sec:hmu_control}.


\subsection{Probabilistic graphical model for joint decision of model parameter and multiplier}\label{sec:pgm}



\begin{figure}
\centering
\includegraphics[scale=0.6]{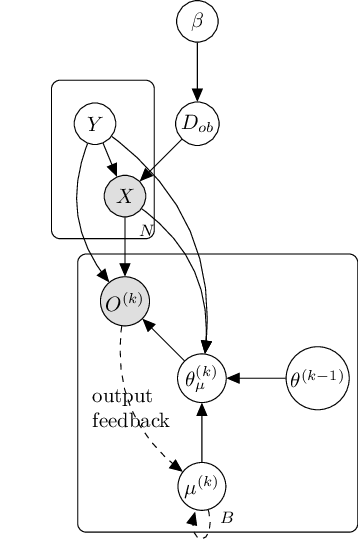}
\caption{Probabilistic graphical model for the sequential decision process of joint model parameter and multiplier adaptation in multi-domain structral risk minimization in~\Cref{eq:srm}. See~\Cref{fig:control_diagram_fbopt} for the control diagram counterpart of the same process.}\label{fig:hyper_para_tune_bayes}
\end{figure}

Gradient based optimization algorithms for neural networks define a dynamic system for model parameters where the weight multiplier in~\Cref{eq:srm} configures such a dynamic system via defining a loss landscape (see~\Cref{fig:landscape_scheduling}). Due to the stochastic nature of the gradient with different subsample and mini-batch distributions for each iteration, it is appropriate to use a probabilistic graphical model~\citep{koller2009probabilistic} to describe such a stochastic process, which has been used to describe sequential decision processes like reinforcement learning~\citep{levine2018reinforcement, sun2019tutorial}. Instead of fixed multipliers, we also model multiplier choice adaptively and propose the probabilistic graphical model in~\Cref{fig:hyper_para_tune_bayes} to describe the sequential decision process of the joint model parameter and multiplier adaptation.

In the upper part of~\Cref{fig:hyper_para_tune_bayes}, we model the generative process of $N$ observations of $X$ (e.g.~an image) and the corresponding supervision signal $Y$ (e.g.~a class label or self-supervision). The path $\beta \rightarrow D_{ob} \rightarrow X$ models domain-specific data generation (information not captured by $Y$), where $D_{ob}$ modeling the data-site~\cite{sun2019high} with hyper-prior $\beta$.

In the lower part of~\Cref{fig:hyper_para_tune_bayes}, with superscript $k$ indexing the iteration of the decision process, we use $\mu^{(k)}$ to represent the multiplier of the loss in \Cref{eq:srm} at the $k$th iteration.
The multiplier $\mu^{(k)}$ serves as a configuration parameter for the optimization process of \Cref{eq:srm} that affects the value of $\theta_{\mu}^{(k)}$ (evolved from its previous value $\theta^{(k-1)}$). 
$\theta_{\mu}^{(k)}$ then becomes the initial value for the next iteration. This can be described in \Cref{eq:controller_theta_abs}. 
The model parameter $\theta^{(k)}$ and observed data $X$, supervision signal $Y$ co-parent the performance indicator $\PerfInd^{(k)}$ as output of the dynamic system. See different realizations of $O$ in~\Cref{remark:pgm_bo_realization} and~\Cref{remark:pgm_bo_fbopt_realization}.

The plate replicates $B$ in the lower part represents the number of optimization iterations (each iteration correspond to $N$ observations).  
The adaptive generation of the next multiplier $\mu^{(k+1)}$ depends on the previous value $\mu^{(k)}$ (the self-loop dashed arrow in \Cref{fig:hyper_para_tune_bayes}) and the feedback information of the output $O^{(k)}$ (long dashed arrow), as in \Cref{eq:controller_mu_abs}.


\begin{equation}
\mu^{(k+1)} = f_{\mu}(\mu^{(k)}, O^{(k)}, \cdot) \label{eq:controller_mu_abs}
\end{equation}
\begin{remark}\label{remark:pgm_bo_realization}
A realization of $f_{\mu}$ can be based on the acquisition function in Bayesian optimization~\citep{garnett2023bayesian} for approximating~\Cref{eq:hyperopt}. See an example described in~\citet{sun2019high}, where $O$ in~\Cref{eq:controller_mu_abs} is chosen to be the validation set prediction performance and iteration index $k$ corresponds to a whole training cycle of \Cref{eq:ml_theta} to iteratively achieve \Cref{eq:hyperopt}. 
\end{remark}

\begin{remark}\label{remark:pgm_bo_fbopt_realization}
Different from the Bayesian optimization realization of this probabilistic graphical model, which changes multipliers at a much slower timescale than that for the evolution of model parameters in neural network training, the hierarchical output feedback optimization we proposed in~\Cref{fig:control_diagram_fbopt} and elaborated in~\Cref{sec:hmu_control} and following sections operates at the timescale of epoch, thus forms another realization of this probabilistic graphical model with the iteration index $k$ corresponding to an epoch. In addition, we have $O(\cdot)=\ell(\cdot), R(\cdot)$ in \Cref{fig:hyper_para_tune_bayes} in contrast to validation set performance used in Bayesian optimization.   
\end{remark}

\begin{figure}[!ht]
\centering
\includegraphics[width=0.5\linewidth]{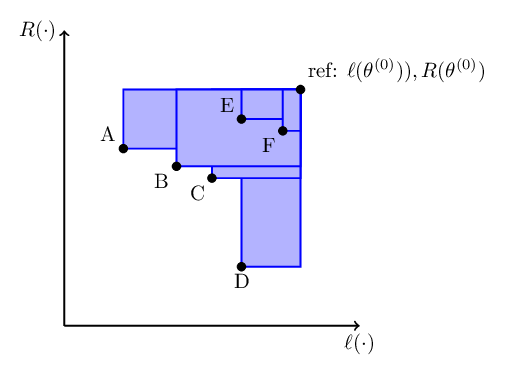}
\caption{Illustration of $\hyperv$ in \Cref{def:hv}:  We take $[\ell(\theta^{(0)}, \cdot), R(\theta^{(0)})]$ as reference point. Suppose $\{A,B,C,D,E,F\}$ in the illustration constitutes function values of the reachable set $\vreachable\left(\theta^{(0)},f_{\theta}(\cdot)\right)$
in~\Cref{def:reachable_set}. 
For any $\theta$ corresponding to the point A in the illustration with coordinate $[\ell(\theta, \cdot), R(\theta, \cdot)]$, $\hyperv$ maps $\theta$ to $\mnondom{\theta}$, then to $\mnondomfront{\theta}$ (see \Cref{def:pareto_ec}), which is the point set \{A,B,C,D\}, then it calculates the dominated hypervolume with respect to the reference point (union of the shaded rectangles). }\label{fig:hypervolume}
\end{figure}

If the joint decision process of model parameter and multiplier is optimal, from a statistical inference point of view (See~\Cref{remark:est_control_dual}), we are conducting a sequential inference (i.e., finding a sequence of $\mu$ in \Cref{eq:mpc_obj}), such that at the last step $B$, the final value $\theta_{\mu^{(B)}}$ has the best possible hyper-volume (in a stochastic sense, see~\Cref{fig:hypervolume}) with respect to output $O$ (see~\Cref{remark:pgm_bo_realization} and~\Cref{remark:pgm_bo_fbopt_realization}).

We introduce an objective function for estimating $\theta_{\mu^{(B)}}$ based on the profile likelihood in \cref{eq:likelihood_hv} in accordance with the probabilistic graphical model in \Cref{fig:hyper_para_tune_bayes}. This likelihood promotes a high $\hyperv(\cdot)$ in \Cref{def:hv} with respect to $\theta_{\mu^{(B)}}$.
$\Xi$ is the normalization factor.
\begin{align}
\mathcal{P}(\theta_{\mu^{(B)}}) & = \frac{1}{\Xi}\exp\left(\hyperv_{
O(\cdot)
}(\theta_{\mu^{(B)}},\theta^{(0)}, f_{\theta},\cdot)\right)\label{eq:likelihood_hv}
\end{align}
\begin{definition}\label{def:hv} 
$\hyperv_{O(\cdot)=\ell(\cdot), R(\cdot)}(\theta, \theta^{(0)}, f_{\theta},\cdot)$ maps $\theta$ to the dominated hypervolume~\citep{zitzler2004indicator,zitzler2007hypervolume,guerreiro2021hypervolume} of $\mnondomfront{\theta}$ in \Cref{def:pareto_ec} with respect to reference point $\ell(\theta^{(0)}, \cdot)\in \mathbb{R}, R(\theta^{(0)}, \cdot) \in \mathbb{R}^d$. As illustrated in \Cref{fig:hypervolume}.
\end{definition}

\begin{remark}
  The probabilistic graphical model in~\Cref{fig:hyper_para_tune_bayes} can also be used to describe other multi-objective deep learning methods~\citep{mahapatra2021exact} via online combinatorial choice of multipliers.    
\end{remark}
\begin{remark}
The probabilistic graphical model also provides a potential possibility to impose a Bayesian interpretation to the joint model parameter training process and multiplier tuning for frequentist machine learning, the deeper investigation of which we leave for future work.
\end{remark}

\begin{remark}[The dual relationship between estimation and control]\label{remark:est_control_dual}
Analogous to how optimal control~\cite{lewis2012optimal} can be formulated as maximum likelihood estimation in a probabilistic graphical model~\citep{levine2018reinforcement,sun2019tutorial}, 
the estimation of $\theta^{(B)}$ based on the profile likelihood like objective in \Cref{eq:likelihood_hv} for \Cref{fig:hyper_para_tune_bayes} is also a sequential decision process and can be solved via optimal control. 
\end{remark}


\subsection{Hierarchical control formulation of joint model parameter and multiplier optimization}\label{sec:hmu_control}
\subsubsection{Optimal control formulation}
Continuing from~\Cref{remark:est_control_dual}, to reformulate the estimation problem as an optimal control problem, we present the sequential decision on choosing multipliers $\mu^{(1)}, \ldots, \mu^{(B)}$ defined in \Cref{eq:srm} with respect to  the objective function in \Cref{eq:mpc_obj} 
at iteration $B$ which promotes optimization of~\Cref{eq:likelihood_hv}.
\begin{align}
\min_{\mu^{(1)}, \ldots, \mu^{(k)}, \ldots, \mu^{(B)}} &\hyperv_{O(\cdot)}(\theta_{\mu^{(B)}},\theta^{(0)}, f_{\theta},\cdot)\label{eq:mpc_obj}\\
\mathrm{s.t.}~\theta^{(k)} &= f_{\theta}(\mu^{(k)}, \theta^{(k-1)}, \cdot)\label{eq:controller_theta_abs}\\
                           O^{(k)}&=\left[\ell(\theta^{(k)} | \ldots), R(\theta^{(k)} | \ldots)\label{eq:output_R_and_ell}\right]\\
                           k&=0, 1, \ldots, B
\end{align}
Here, we use $\theta$ to denote the neural network weights, which can be regarded as the state of a dynamic system, where the state transition is governed by a low level model parameter optimization dynamic $f_{\theta}$ in \Cref{eq:controller_theta_abs} (e.g.~\citet{kingma2014adam}). From a control theory point of view, $f_{\theta}$ depicts the uncertain dynamic of a plant to be controlled, while the task here is to define a controller to generate a $\mu$ sequence to guide the uncertain dynamic of $\theta$ with respect to plant output $O$, as depicted in \Cref{fig:control_diagram_fbopt}. We treat the loss terms $R(\cdot)$ and $\ell(\cdot)$ as the output of the system in \Cref{eq:output_R_and_ell}.
From a statistical point of view, $f_{\theta}$ corresponds to the parent structure of $\theta^{(k)}$ in \Cref{fig:hyper_para_tune_bayes}, which drives $\theta$ to the next value. 

\begin{remark}[closed loop dynamic of model parameter]\label{remark:closed_loop}
A concrete form of \Cref{eq:controller_mu_abs} will detail how $\mu$ is updated by a controller $f_{\mu}$. 
Combining \Cref{eq:controller_mu_abs,eq:controller_theta_abs}, we have 
\begin{align}
\theta^{(k+1)} &= f_{\theta}(f_{\mu}(\mu^{(k)}, O^{(k)}),\theta^{(k)}, \cdot)\label{eq:theta_mu_dyn}\\
&=f_{\theta, \mu}^{k}(\mu^{(0)},\theta^{(0)}, \cdot)
\label{eq:theta_closed_loop}
\end{align}
where we use $f_{\theta,\mu}^{k}$ to represent the $k$-times compound recursive evaluations of $f_{\theta}$ and $f_{\mu}$ in \Cref{eq:theta_mu_dyn} until the dependence is only on $\theta^{(0)},\mu^{(0)}$.
We call \Cref{eq:theta_closed_loop} the \emph{closed-loop} dynamics for $\theta$. 
\end{remark}

Although we have no complete knowledge of the behavior of the low level optimization algorithms described by $f_{\theta}$ in~\Cref{eq:controller_theta_abs}, the penalized loss from~\Cref{eq:srm} often descends after some iterations (possibly with oscillations). This leads to~\Cref{def:ploss_descent_operator}. 
\begin{definition}[Penalized loss descent operator] A penalized loss descent operator $\mathcal{G}$ satisfies that if
$\theta^{+}\in \mathcal{G}_{\mu}(\theta; \ell(\cdot), R(\cdot))$, then $\ell(\theta^{+})+\mu R(\theta^{+})<\ell(\theta)+\mu R(\theta)$\label{def:ploss_descent_operator}\end{definition}
\begin{remark}[Penalized Loss Descent Assumption]\label{remark:ploss_descent_operator}
We take the frequent existence and occurrences of such operators defined in~\Cref{def:ploss_descent_operator} during neural network training as a mild assumption in our following discussion. 
For instance, we could run $f_{\theta}$ in \Cref{eq:controller_theta_abs} in one step to have $\ell(\theta^{(k+1)})+\mu^{(k+1)} R(\theta^{(k+1)})<\ell(\theta^{(k)})+\mu R(\theta^{(k)})$, or in more than one steps to achieve \Cref{def:ploss_descent_operator}. 
\end{remark}

Our final goal, however, is to ensure a joint descent of $\ell(\cdot)$ and $R(\cdot)$, which leads to~\Cref{def:pareto-descent-operator}.
\begin{definition}\label{def:pareto-descent-operator}
A Pareto-descent operator takes $\theta$ to $\theta^{(+)}$ (via one or several iterations) which descents $R(\cdot)$ and $\ell(\cdot)$ simultaneously, i.e.,
\begin{equation}
\ell(\theta^+) < \ell(\theta) \quad \text{and} \quad R(\theta^+) \preceq R(\theta) \label{eq:combined}
\end{equation}
\end{definition}
\subsubsection{Multi-objective optimization via shrinking the reference signal in constrained optimization}\label{sec:constrained_opt_shrink_reference}
How can we design a closed-loop dynamic system in~\Cref{remark:closed_loop} to ensure a Pareto descent of $O(\cdot)$, leading to multi-objective optimization in \Cref{def:pareto-descent-operator} or \Cref{eq:mpc_obj}?
To resolve this, we reduce it to a simpler problem, i.e., when bounding the regularization term $R(\cdot)$ by a time-varying reference bound $b^{(k)}$ (a.k.a.~setpoint) in~\Cref{eq:feasibility_r}, what value can we achieve for $\ell(\cdot)$ at~\Cref{eq:main_obj}. If we could ensure the reference bound $b^{(k)}$ changes monotonically, multi-objective optimization can be achieved. 

Thus, we define the multi-dimensional reference bound $b^{(k)}$ (setpoint) in~\Crefrange{eq:setpoint_ini}{eq:setpoint_ada_abs}:
\begin{align}
b^{(0)}&=\rho R^{(0)}, 0<\rho\in \mathbb{R} < 1\label{eq:setpoint_ini}\\
b^{(k)}&=g_b(\ell^{(0:k)}(\cdot), R^{(0:k)}(\cdot))\label{eq:setpoint_ada_abs}
\end{align}
where $g_b({\cdot})$ represents the mapping from the previous value at the training step $k-1$ to the new value at step $k$, defined as~\Cref{eq:b_pareto_descent}.
\begin{subnumcases}{b^{(k)}=}
R^{(k)},~\text{if } R^{(k)} \prec b^{(k-1)}~\textbf{and}\nonumber\\ 
\ell^{(k)}<\min_{j=0,\cdots, k-1} \ell^{(j)},\label{eq:b_pareto_descent}\\
b^{(k-1)},~\text{otherwise}
\end{subnumcases}

\begin{remark}\label{remark:b_shrink_incidate_pareto_descent}
The shrinkage of $b^{(k)}$ in~\Cref{eq:b_pareto_descent} also depends on $\ell(\cdot)$ decrease, thus indicates multi-objective Pareto-descent in \Cref{def:pareto-descent-operator}.    
\end{remark}

\begin{restatable}[Pareto-descent via constrained optimization with shrinking bound]{prop}{LabelRestatebleParetoDescentViaConstraintOptShrinkBound}\label{prop:pareto_descent_constrained}
With $b^{(k)}$ defined in~\Cref{eq:b_pareto_descent}, suppose the following constrained optimization~\cite{bertsekas2014constrained} problem in~\Crefrange{eq:main_obj}{eq:feasibility_r} starting with $\theta^{(k)}$, under $s_k+m_k$ number of iterations ($s_k>0$ and $m_k\ge 0$) has a solution 
\begin{align}
&\min_{\mu^{(k+1)}, \ldots, \mu^{(k+s_k+m_k)}}~\ell^{(k+s_k+m_k)}(\cdot)\label{eq:main_obj}\\ 
\quad\mathrm{s.t.~}
R^{(j_1)}(\cdot) &\nprec b^{(k)}, j_1=k, \ldots, k+s_k-1\label{eq:not_feasibility_r}\\
\theta^{(j+1)} &= f_{\theta}(\mu^{(j+1)}, \theta^{(j)}, \cdot)~(\Cref{eq:controller_theta_abs})
\nonumber\\
R^{(k+j_2)}(\cdot) &\prec b^{(k)}, j_2=s_k, \ldots, s_k+m_k\label{eq:feasibility_r}
\end{align}
we achieve multi-objective descent in \Cref{def:pareto-descent-operator} at step $k+s_k+m_k$ compared to step $k$.
\end{restatable}
\begin{proof}
See~\Cref{proof:constraint_opt_with_bound_shrink_constraint_lead2_Pareto}.
\end{proof}

\begin{restatable}[Approximation of constraint optimization]{corollary}{LabelRestatebleParetoDescentViaConstraintOptShrinkBoundCoro}\label{coro:pareto_descent_constrained}
Regardless of the attainability of the minimization in \Cref{eq:main_obj}, as long as we have 
\begin{align}
\ell^{(k+s_k+m_k)}(\cdot) < \ell^{(k)}(\cdot)\label{eq:l_decrease} 
\end{align} 
we achieve multi-objective descent at step $k+s_k+m_k$. 
\end{restatable}
\begin{proof}
  See~\Cref{proof:coro_constraint_opt_with_bound_shrink_constraint_lead2_Pareto}.
\end{proof}

We discuss how to approximate the constrained optimization in~\Cref{sec:loss_landscape_scheduling}. 
\subsubsection{Loss landscape scheduling}\label{sec:loss_landscape_scheduling}
\begin{figure}[htbp]
\centering
\includegraphics[width=0.6\linewidth]{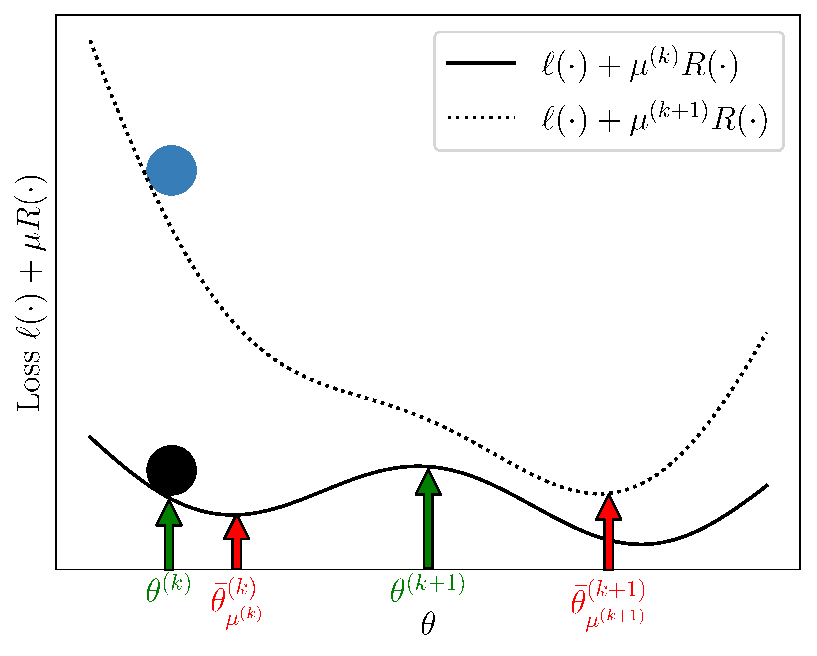}
\caption{Illustration of multiplier induced loss landscape scheduling, used as control signal in \Cref{fig:control_diagram_fbopt}: The lifted landscape $\ell(\cdot)+\mu^{(k+1)}R(\cdot)$ (dotted curve) scheduled at iteration $k+1$, 
  enables the model parameter dynamic to overcome the local minimum $\bar{\theta}^{(k)}_{\mu^{(k)}}$ of the old loss landscape $\ell(\cdot)+\mu^{(k)}R(\cdot)$ (solid curve) scheduled at iteration $k$, as can be imagined via the two balls with different colors rolling along the corresponding loss landscapes. 
This showcases a scenario corresponding to \Cref{def:reg-pareto-slider}: In comparison to $\theta^{(k)}$, $\theta^{(k+1)}$ corresponds to a decreased $\ell(\cdot)+\mu^{(k+1)}R(\cdot)$ value but increased $\ell(\cdot)+\mu^{(k)}R(\cdot)$ value. }\label{fig:landscape_scheduling}
\end{figure}

The evolution of the model parameters can be regarded as a plant with uncertain dynamic $f_{\theta}$ in \Cref{eq:controller_theta_abs}, where the only control we have at hand is the sequence $\mu^{(k)}$, which provides different loss landscapes at different iterations 
as depicted in \Cref{fig:landscape_scheduling}.   
How can we design such 
loss landscape sequence to promote \Cref{eq:feasibility_r} to happen? 
To simplify the discussion, first consider $\mu, R(\cdot)\in \mathbb{R}_{+}^{d=1}$. Before the constraint in \Cref{{eq:feasibility_r}} is satisfied, design $\mu^{(k+1)},\ldots,\mu^{(k+s_k)}$ to be an increasing sequence~\cite{bazaraa2013nonlinear}, which gives more weight to the corresponding regularization term in $R(\cdot)$, to approximate the constraint optimization in \Cref{eq:main_obj} by optimizing \Cref{eq:srm} for each $\mu$ until \Cref{eq:feasibility_r} is satisfied. However, will the increased weight on the regularization term in $R(\cdot)$ result in deteriorated $\ell(\cdot)$?
One possibility can be, although $f_{\theta}(\mu^{(k+1)},\theta^{(k)}, \cdot)$ in~\Cref{eq:controller_theta_abs} brings $\theta^{(k)}$ to the next value $\theta^{(k+1)}$ which decreases the penalized loss in \Cref{eq:srm} with $\mu^{(k+1)}$, $\theta^{(k+1)}$ evaluates to a deteriorated penalized loss with respect to the old $\mu^{(k)}$ as shown in \Cref{fig:landscape_scheduling}. To describe this behavior during the increasing process of $\mu$, we have \Cref{eq:deteriorate_old_lambda} in \Cref{def:reg-pareto-slider}. This situation is of particular interest due to the possibility of overcoming local minima of a fixed loss landscape, as indicated in \Cref{fig:landscape_scheduling}.

\begin{definition}\label{def:reg-pareto-slider} 
A reg-Pareto slider $\mathcal{A}(\cdot)$ with respect to function $\ell(\cdot)$ and $R(\cdot)$ is defined to be a set map, s.t.~if
$\{\theta^{(k+1)}, \mu^{(k+1)}\} \in \mathcal{A}(\mathcal{\ell}(\cdot), R(\cdot);\theta^{(k)}, \mu^{(k)})$ then
{\scriptsize
\begin{align}
\ell(\theta^{(k+1)}) + \mu^{(k+1)} R(\theta^{(k+1)})&\le \ell({\theta}^{(k)}) + \mu^{(k+1)} R({\theta}^{(k)})\label{eq:improve_new_lambda}\\
\ell(\theta^{(k+1)}) + \mu^{(k)} R(\theta^
{(k+1)})&\ge \ell({\theta}^{(k)}) + \mu^{(k)} R({\theta}^{(k)})\label{eq:deteriorate_old_lambda}\\
\mu^{(k+1)}&\ge \mu^{(k)} \label{eq:mu_increase}
\end{align}}
\end{definition}

\subsubsection{Analysis}\label{sec:analysis}
Based on~\Cref{def:reg-pareto-slider}, we have the following conclusion in~\Cref{corollary:reg-slide} where we show the increase of $\ell(\cdot)$ as an expense to decrease of $R(\cdot)$ is bounded. 
\begin{restatable}[Bounded Pareto Trade-off]{prop}{LabelRestatebleRegSlide}\label{corollary:reg-slide}
Let
\begin{align}
\{\theta^{(k+1)},  \mu^{(k+1)}\} \in \mathcal{A}(\mathcal{\ell}(\cdot), R(\cdot);\theta^{(k)}, \mu^{(k)}) 
\end{align} from \cref{def:reg-pareto-slider}
assume $R(\cdot)\in \mathbb{R}^{d=1}$, then
{\scriptsize
\begin{align}
R(\theta^{(k+1)})&\le R({\theta}^{(k)})\\
\ell(\theta^{(k+1)})&\ge \ell({\theta}^{(k)}) \\
\ell(\theta^{(k+1)})-\ell({\theta}^{(k)})&\le \mu^{(k+1)}( R({\theta}^{(k)})  -  R(\theta^{(k+1)}))\label{eq:l_increase_bound}
\end{align}}
\end{restatable}
\begin{proof}
See~\Cref{proof:reg_slider}.
\end{proof}

\begin{remark}[Multi-dimensional $R$]\label{remark:vector_pareto_slider}
In the case of $R(\cdot)\in \mathbb{R}^{d>1}$, when all components of $\mu$ increase, \Cref{eq:improve_new_lambda} and \Cref{eq:deteriorate_old_lambda} imply at least one component of $R(\cdot)$ will decrease. We defer to \Cref{remark:decrese_mu_overshoot} and \Cref{sec:pid} for the discussion on when some components of $\mu$ increase while other components decrease. 
\end{remark}

\begin{remark}[Decrease of $\ell(\cdot)$]\label{remark:other_situation_l_r_decrease}
\Cref{corollary:reg-slide} give a condition of decreasing $R(\cdot)$ at the expense of increasing $\ell(\cdot)$ (bounded though by~\Cref{eq:l_increase_bound}) at iteration $k$. When the gradient component from $\ell(\cdot)$ and $R(\cdot)$ agrees with each other at another iteration, we could expect both objectives to decrease as in \Cref{def:pareto-descent-operator}. It can also be the case at a particular iteration, $\ell(\cdot)$ decreases at the expense of $R(\cdot)$ increases.\end{remark}  
What is left unclear, however, is after the accumulative effect of several iterations, whether these exists an $s_k,m_k$ ensuring
\Cref{eq:l_decrease}. 
To answer this,
we derive~\Cref{corollary:ellbound}, which 
gives a bound on the accumulative change of $\ell(\cdot)$  with $B=s_k+m_k$. 
\begin{restatable}[Single-step $\ell$ increase bound]{prop}{LabelRestatebleEllBoundTelescope}\label{corollary:ellbound}
    Suppose that for all $k=0,\dots,B-1$
    \begin{align*} 
        \ell(\theta^{(k+1)}) + \mu^{(k+1)} R(\theta^{(k+1)}) &\le \ell(\theta^{(k)}) + \mu^{(k+1)} R(\theta^{(k)})
    \end{align*}
    and $\mu^{(k+1)} > \mu^{(k)}$. Then
    \begin{align}\label{eq:ell_k_bound}
        \ell(\theta^{(B)}) \le \ell(\theta^{(0)}) + S_> + S_<,
    \end{align}
    where 
    \begin{align*}
        S_> &= \sum_{k \in \mathcal{K}_>} \mu^{(k+1)} (R(\theta^{(k)}) - R(\theta^{(k+1)})), \\
        S_< &= \sum_{k \in \mathcal{K}_<} \mu^{(k+1)} (R(\theta^{(k)}) - R(\theta^{(k+1)})),\\
        \mathcal{K}_> &= \{k \in \{0,\dots,B-1\}:~\Cref{eq:deteriorate_old_lambda}~\text{holds.}\}, \\
        \mathcal{K}_< &= \{0,\dots,B-1\} \setminus \mathcal{K}_>.
    \end{align*}
\end{restatable}
\begin{proof}
See~\Cref{proof:reg_ell_bound}.
\end{proof}

Based on~\Cref{corollary:ellbound}, we have the following conjecture to ensure \Cref{eq:l_decrease}
\begin{conj}[Multi-step $\ell$ decrease]\label{conj:l_decrease}
Further decompose $S_{<}$ from \Cref{corollary:ellbound} into $S_{<}^{-}$ and $S_{<}^{+}$ via decomposing $\mathcal{K}_<$ in \Cref{corollary:ellbound} into $\mathcal{K}_{<}^{-}$ and $\mathcal{K}_{<}^{+}$. Here $\mathcal{K}_{<}^{-}$ corresponds to the situations of $\ell(\cdot)$ descent with $S_{<}^{-}<0$. And $\mathcal{K}_{<}^{+}$ corresponds to the $\ell(\cdot)$ ascent cases but upperbounded by $S_{<}^{+}=\sum_{k \in \mathcal{K}_<^+} \mu^{(k+1)} (R(\theta^{(k)}) - R(\theta^{(k+1)}))$. Then $\exists$ sequence $\{\mu^{(k)}\}$ s.t. 
\begin{align}\label{eq:ell_descent_conj}
\ell(\theta^{(B)}) - \ell(\theta^{(0)}) \le S_> + S_<^{+}+S_<^{-}<0
    \end{align}
\end{conj}

\begin{remark}[Trade-off Multi-dimensional $\mu$]\label{remark:decrese_mu_overshoot}
In the case of $d>1$, as long as one loss component of ${R}^{(k)}(\cdot)$, without loss of generality, say $R_{1}^{(k)}(\cdot)$ is not bounded by its corresponding setpoint component $b_1^{(k)}(\cdot)$, we can still increase the value of the $\mu$ component $\mu^{(k)}_1$ to give more weight to the corresponding loss component, until the loss component in question decreases below the setpoint. However, an increased $\mu$ component $\mu^{(k)}_1$ corresponds to more weight on the gradient component corresponding to $R_{1}^{(k)}(\cdot)$, which makes it harder for other components of $R(\cdot)$ and $\ell(\cdot)$ to decrease. Therefore, instead of always increasing each component of $\mu$, if one component of $R$ has overshoot over the setpoint (i.e.~constraint switched from being not satisfied to satisfied with respect to a particular loss component), we decrease the $\mu$ value to give the other components of $\mu$ and $R$ more feasible space to adjust themselves.\end{remark}
\begin{remark}[Multi-step Pareto Descent]\label{remark:multi_step_pareto_descent}
We could extend~\Cref{conj:l_decrease} to $\exists \mu$ sequence such that after $B$ iterations, Pareto descent in \Cref{def:pareto-descent-operator} could be achieved (equivalently, setpoint should shrink see \Cref{remark:b_shrink_incidate_pareto_descent}). We did observe such a Pareto descent behavior in our experiment for the $d>1$ case of \Cref{eq:loss_diva_srm}, e.g.~see \Cref{fig:dyn_pacs_diva_fbopt} and \Cref{fig:phase_portrait_pacs_diva_fbopt}.    
\end{remark}

\begin{algorithm}
\caption{\fontsize{5}{6}\selectfont 
{Multi-objective hierarchical output feedback optimization}}\label{alg:fbopt}
\begin{algorithmic}[1]
\Procedure{M-Hof-Opt}{$\mu^{(0)},\theta^{(0)}, \rho, \cdot$}
\State Initialize $\mu^{(0)}$, calculate $b^{(0)}=\rho R^{(0)}$. 
\State Compute $K_I$ based on Remark~\ref{remark:choiceK}.
\While{budget $B$ not reached}
\State Update $\mu$ from controller in \Cref{eq:pid}.
\State Update $\theta$ via~\Cref{eq:controller_theta_abs} with updated $\mu$.
\State Adapt setpoint according to \Cref{eq:setpoint_ada_abs}.  
\EndWhile\label{algo:fbopt_while}
\State \textbf{return} $\theta^{(B)}$ \Comment{}
\EndProcedure
\end{algorithmic}
\end{algorithm}


\subsubsection{Output feedback PI-like controller}\label{sec:pid}
Based on~\Cref{remark:decrese_mu_overshoot},
with the adaptive law from \Cref{eq:setpoint_ada_abs,eq:b_pareto_descent} for setpoint $\setpoint$, we design the following controller (see~\Cref{fig:control_diagram_fbopt}) for $\mu$:  
\begin{align}
e^{(k)} &= R(\cdot)- b^{(k)}\label{eq:delta}\\
\delta_I^{(k+1)} &= (1-\xi_d) \delta_I^{(k)}+\xi_d e^{(k)}\label{eq:delta_ma}\\
\mu^{(k+1)} & =\max( \mu^{(k)}\exp^{\max(K_{I}\delta_I^{(k+1)}, v_{sat})}, \mu_{clip})
\label{eq:pid}\\
K_I&>0\\
K_I&\in \mathbb{R}^d\\
k&=0, 1, \ldots, B
\end{align}
In \Cref{eq:delta}, we calculate how far the current output is away from the setpoint $\setpoint$, which gets passed through a moving average in \Cref{eq:delta_ma} with $\xi_d$ being the coefficient of moving average in \Cref{eq:delta_ma}~\citep{rezende2018taming}.
$\mu$ is the output of the controller in \Cref{eq:pid}, $K_I$ is the control gain for PI (proportional-integration)~\citep{johnson2005pid} like control, $K_I\delta_I^{(k+1)}$ is component wise multiplication, $\exp$ and $\max$ are computed component wise, $v_{sat}$ is the exponential shoulder saturation, which defines the maximum rate of change of multipliers. $\mu_{clip}$ is the upper bound for $\mu$. Note that $\mu_{clip}$ determines the dynamic range of $\mu$ while $v_{sat}$ determines an upper bound about how fast $\mu$ changes.


\begin{remark}\label{remark:choiceK}
To avoid arbitrary choice of $K_I\in \mathbb{R}^d$, we use $\delta^{(0)}=R^{(0)}-b^{(0)}$ in \Cref{eq:delta_ma} where $b^{(0)}=\rho R^{(0)}$ (a percentage $\rho$). We divide by $\eta v_{sat}$ with $0<\eta<1$ to get the value for $K_I$ so that the hyperparameter for our algorithm is $\rho, \eta \in \mathbb{R}$ instead of $K_I\in \mathbb{R}^d$.
\end{remark}

The whole process is summarized in \Cref{alg:fbopt}, which details \Cref{fig:control_diagram_fbopt}.

\subsubsection{Multi-objective setpoint based model selection}\label{sec:msel_setpoint}
In general, the feasibility in \Cref{eq:feasibility_r} becomes more difficult to be met each time the setpoint $b$ shrinks. After several setpoint shrinkages, the difficulty of attaining the feasibility can lead to an oscillating behavior of $R(\cdot)$ around the setpoint with uncontrolled amplitude, as shown in \Cref{fig:dyn_pacs_diva_fbopt}. To circumvent this behavior, the up-to-event best model is selected at the last setpoint shrinkage. 

\subsection{Related work and discussion}
We discuss the advantages of our method compared to other multiplier scheduling and multi-objective optimization methods in~\Cref{sec:related_work}.
\section{Experiments}\label{sec:exp}
\begin{figure*}[ht!]
\centering
\includegraphics[width=0.27\linewidth]{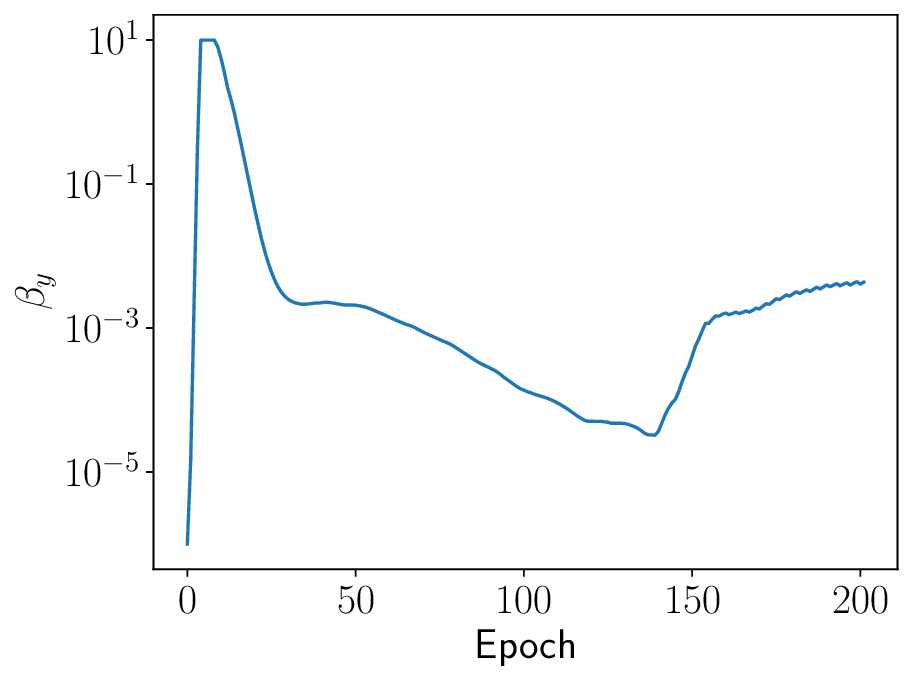}
\includegraphics[width=0.3\linewidth]{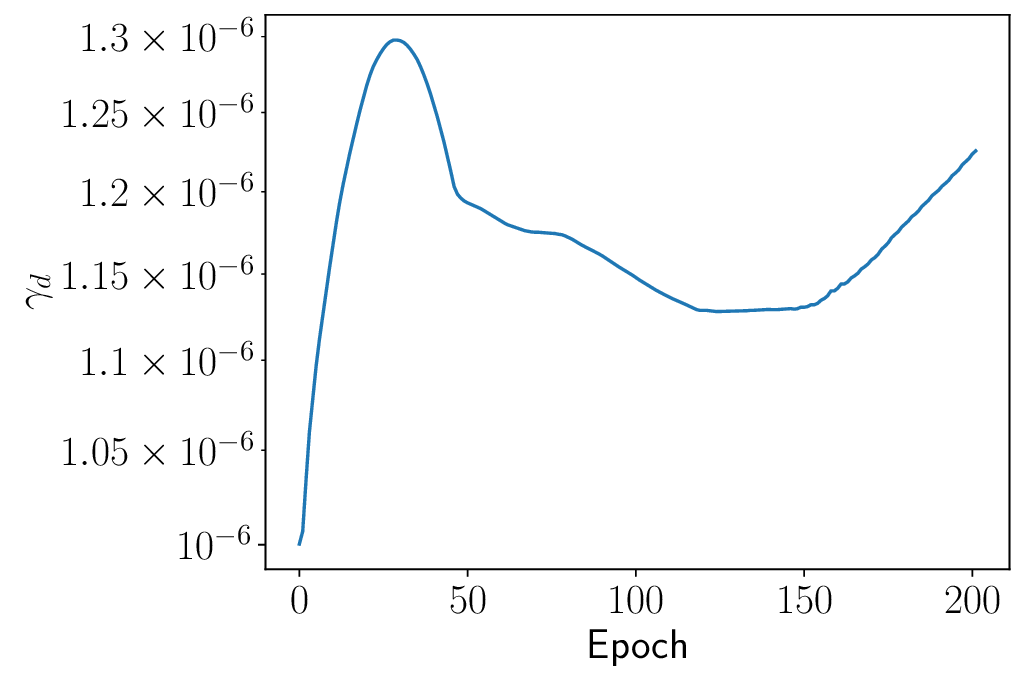}
\includegraphics[width=0.27\linewidth]{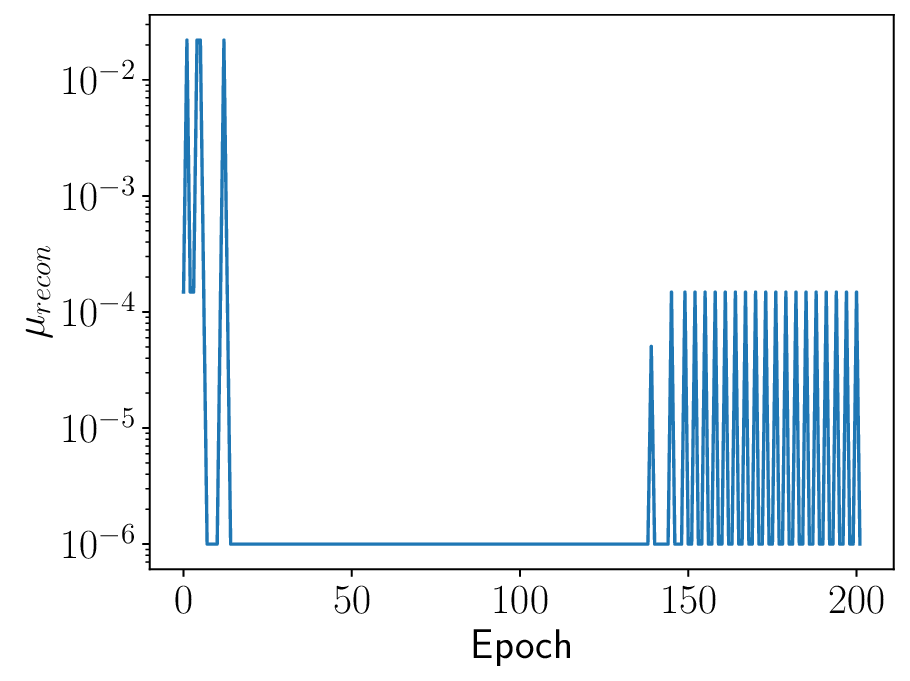}\\
\includegraphics[width=0.27\linewidth]{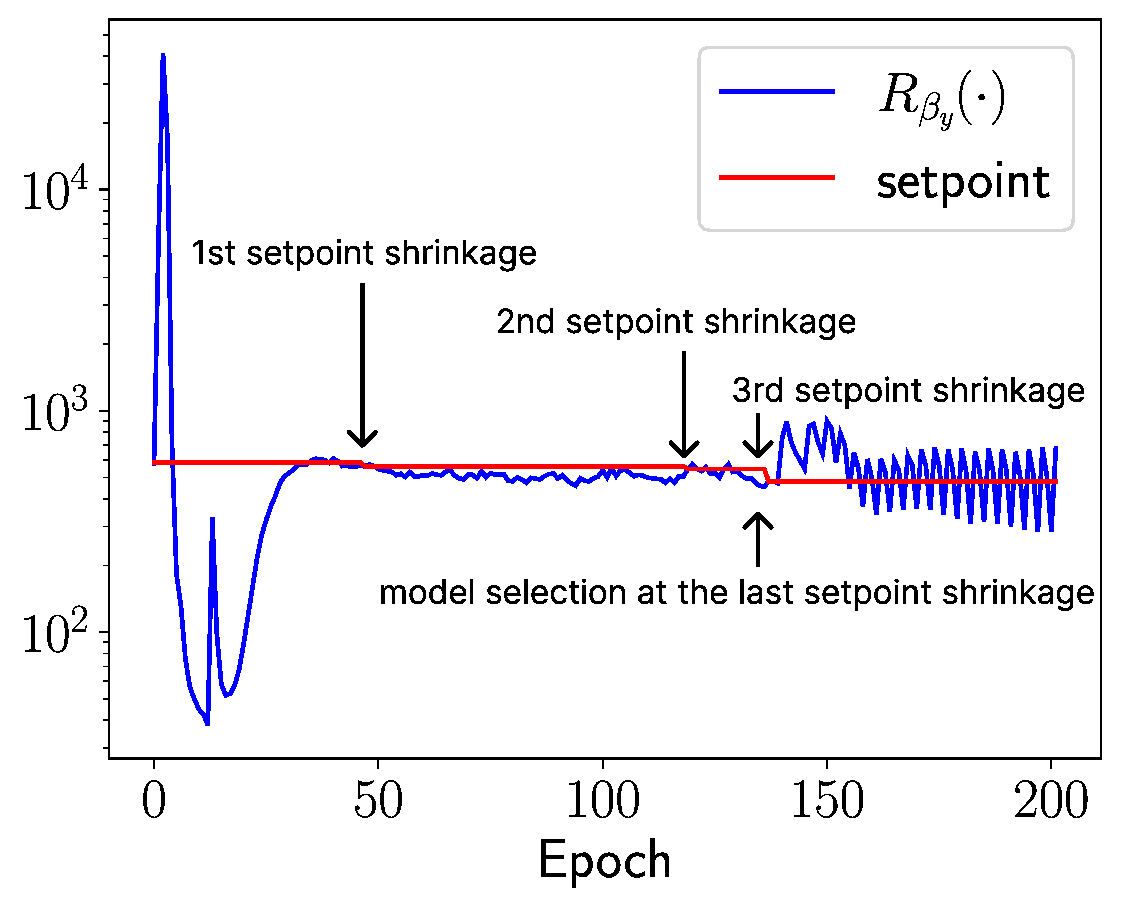}
\includegraphics[width=0.29\linewidth]{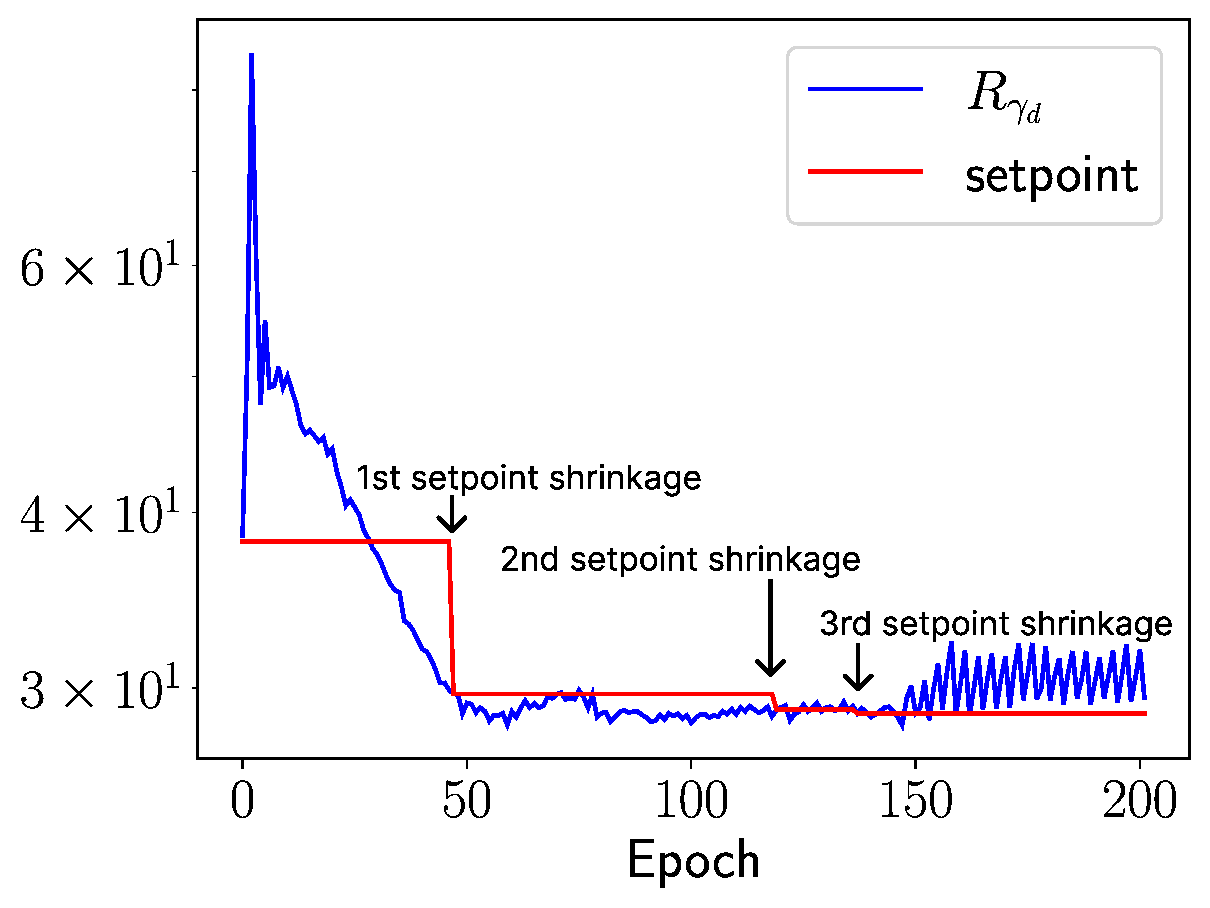}
\includegraphics[width=0.3\linewidth]{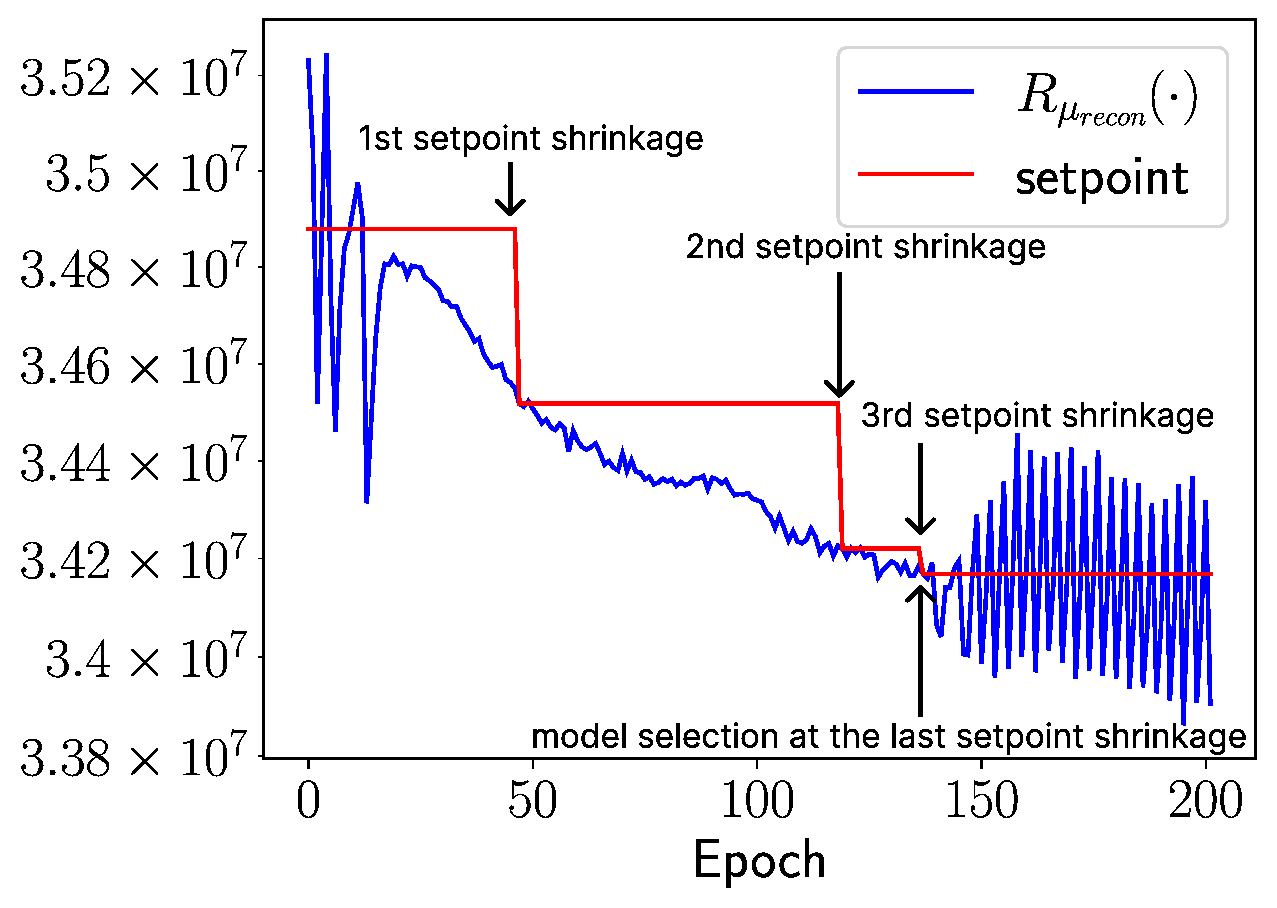}
\caption{Our method drives different loss terms of $R(\cdot)$ at different scales and rates towards the setpoint, which further promotes the setpoint shrinkage. 
In the \textbf{Top row}, we show the multiplier dynamic as controller output signal in \Cref{eq:pid} across training epochs. In the \textbf{Bottom row}, we present
the tracking behavior of the corresponding regularization loss $R(\cdot)$ in \Cref{eq:loss_diva_srm} with respect to setpoint $b$ defined in \Cref{eq:setpoint_ada_abs}. 
}
\label{fig:dyn_pacs_diva_fbopt}
\end{figure*}

We conduct experiments and benchmarks to answer the following questions: Is M-HOF-Opt capable of adjusting multidimensional multipliers automatically to drive each component of $R(\cdot)$ at different scales to the setpoint? Does the setpoint shrink, which implies multi-objective descent? See~\Cref{fig:dyn_pacs_diva_fbopt} and~\Cref{fig:phase_portrait_pacs_diva_fbopt} in~\Cref{subsec:exp_dyn}. 

Will wrong multiplier choices for warmup and fixed multiplier training schemes result in catastrophic effects? In contrast, does varying controller hyperparameters have detrimental effects on M-HOF-Opt? 
See~\Cref{fig:benchmark_pacs_diva} in~\Cref{subsec:exp_benchmark_rsts}, as well as~\Cref{fig:bench_irm_dial} in~\Cref{sec:irm_dial_bench}. 

Does changing the lower-level optimization algorithm (i.e.~the open-loop plant or the feedback-free model parameter dynamic system M-HOF-Opt has to control) have an impact on the performance of M-HOF-Opt? See~\Cref{sec:adamw_cosineLR}. 

With respect to the above questions, we got favorable results for M-HOF-Opt in all experiments which we analyze below. The experimental and benchmark setting is detailed in~\Cref{subsec:exp_setting} with implementation in~\url{https://github.com/marrlab/DomainLab/tree/mhof} from \textit{DomainLab}~\citep{sun2024domainlab}.

\subsection{Illustration of multi-objective descent}\label{subsec:exp_dyn}
Since the domain generalization model DIVA~\citep{ilse2020diva} with math formulation explained in~\Cref{subsec:math_diva} has 6 loss terms\footnote{At time of writing, we are not aware of any other deep learning model with more loss terms.}, it is a good example to demonstrate the advantage of our algorithm in trading off many loss terms. Our hierarchical output feedback optimization training scheme produced the dynamics of $R(\cdot)$ shown in \Cref{fig:dyn_pacs_diva_fbopt} together with the corresponding setpoints $\setpoint$ and multipliers.  
Note that the multiplier $\beta_y, \gamma_d, \mu_{recon}$ and their corresponding loss components operate at different numerical ranges, but our controller still manages to drive the different $R(\cdot)$ loss terms down at different scales and rates. 

\begin{remark}[setpoint shrinkage]\label{remark:setpoint_shrinkage}
Regarding the dynamics of how the setpoint $b$ defined in~\Crefrange{eq:setpoint_ini}{eq:setpoint_ada_abs} decreases in~\Cref{fig:dyn_pacs_diva_fbopt}: Initially, \Cref{eq:not_feasibility_r} holds, in which case we do not update the setpoint, even if some but not all the blue curves (corresponding to $R(\cdot)$ components) are below the red curves (components of the setpoint). After all constraints for each component of $R(\cdot)$ are satisfied as described in \Cref{eq:feasibility_r}, the setpoint (red curves) gets adapted to a new value (which we term shrinkage), as a new goal to be reached by the $R(\cdot)$ loss in \Cref{eq:loss_diva_gamma_d}. 
Note that the shrinkage of setpoint depends on Pareto dominance in \Cref{eq:b_pareto_descent}, while non-dominance defined an equivalent relation in \Cref{def:non_dominance}. 
\end{remark}

\begin{figure}[!ht]
\centering
\includegraphics[width=0.5\linewidth]{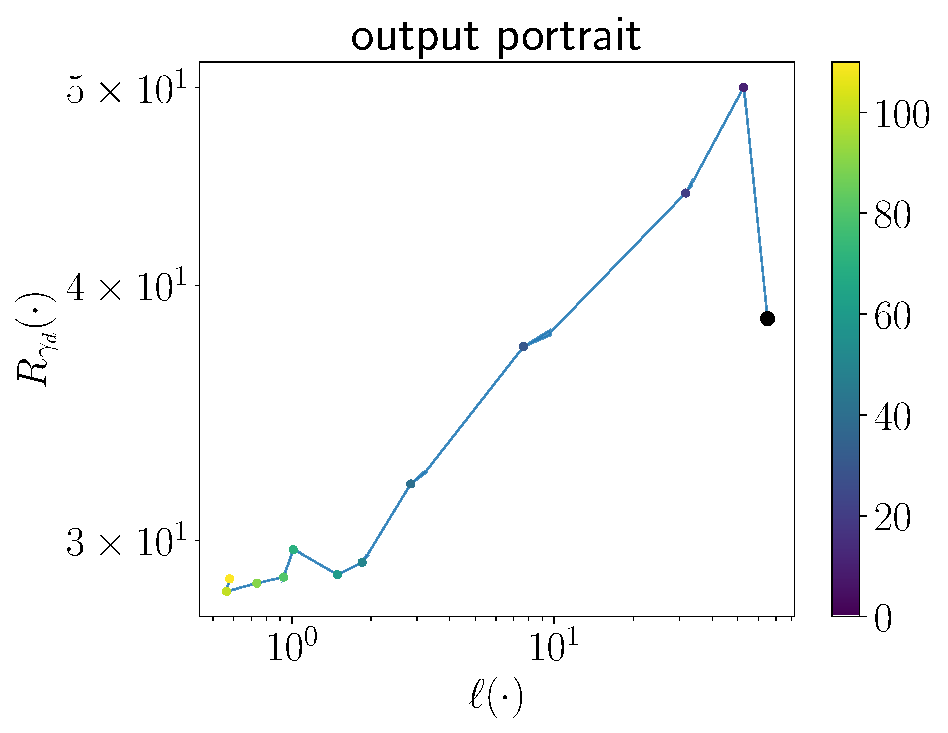}\caption{Our training scheme adapts the multi-dimensional multiplier $\mu$, steering the optimization process towards a configuration that balances the trade-off between minimizing output $\ell(\cdot) = \mathbb{E}_{q_{\phi_y}(z_y|x)}[\log q_{\omega_y}(y|z_y)]$ in \Cref{eq:loss_diva_gamma_y} and 
output $R_{\gamma_d}(\cdot)=\mathbb{E}_{q_{\phi_d}(z_d|x)}[\log q_{\omega_d}(d|z_d)]$ in~\Cref{eq:loss_diva_gamma_d}.
In this plot, the solid large point corresponds to the initial output. We use gradually changing colors to indicate the training iterations (epochs), indicated by the color bar. 
For improved visualization, we exclusively utilized the initial 120 epochs and plotted data points solely for every 10 epochs. This figure corresponds to the experimental setting in~\Cref{fig:dyn_pacs_diva_fbopt}.
}\label{fig:phase_portrait_pacs_diva_fbopt}
\end{figure}

\Cref{fig:phase_portrait_pacs_diva_fbopt} shows the output portrait of $\ell(\cdot)$ versus a component of $R(\cdot)$ of our training scheme 
This confirms the multi-step Pareto descent discussed in~\Cref{remark:multi_step_pareto_descent}.

With the multi-objective setpoint based model selection criteria in \Cref{sec:msel_setpoint}, our method admits a multi-objective descent of the selected model compared to the initial output, such that the uncontrolled behavior at the end of iterations can be safely ignored.

\subsection{Benchmark results}\label{subsec:exp_benchmark_rsts}
We demonstrate the power of M-HOF-Opt in automatic adaptation of multipliers via benchmarking out-of-domain generalization performance in~\Cref{fig:benchmark_pacs_diva} for training DIVA~\cite{ilse2020diva} compared to baseline training schemes. 
From \Cref{fig:benchmark_pacs_diva}, we can observe that the generalization performance of 
baselines  
is highly sensitive to multiplier combination. 
In contrast, our method, as supported by the adaptive behaviors visualized in \Cref{fig:dyn_pacs_diva_fbopt}, automatically adjusts the multipliers during the training process, leading to a robust performance with respect to changes of controller hyperparameters and multiplier initial condition $\mu^{(0)}$. Our method does not need hyperparameter searching of the multiplier, thus saves a fair amount of computational resources. For additional experiments, see~\Cref{sec:irm_dial_bench}.


\section{Conclusion}
This work uses control theory to address the issue of combinatorial choice for multidimensional multipliers weighting many loss terms in structural risk minimization for deep neural networks by proposing a novel multi-objective optimization algorithm via constraint optimization with shrinking bound.
We develop a probabilistic graphical model for joint model parameter and multiplier optimization with respect to a multi-objective descent of all loss terms. The estimation of model parameter leads to an automatic adjustment scheme for the multipliers, adopting a hierarchical control scheme, which breaks the multi-objective optimization problem into a series of constraint optimization sub-problems. Each sub-problem is configured with a self-adaptive multi-objective setpoint updated via Pareto dominance. A PI-like multiplier controller drives the loss term to satisfy the setpoint constraint for each sub-problem. 
Our method operates at the timescale of epoch level during training, thus circumvents the need for exhaustive multiplier search and saves tremendous computational resources compared to methods like Bayesian optimization. It also circumvents the excessive memory requirements and heavy computational burden of existing multi-objective deep learning methods.  
Our method demonstrates robust out-of-domain generalization performance against controller hyperparameter variation compared to other multiplier choice or scheduling schemes which produce very unstable behavior in the training due to the need for a combinatorial choice of multipliers. 

\subsubsection*{Acknowledgements} 
We sincerely thank Dr.~Changxin Liu and Dr.~Prof.~Jann Rolfes for spending hours on proofreading the manuscript and offering many helpful suggestions to the manuscript. We thank Tomas Chobola for \textit{Inkscape} support. We thank the anonymous reviewers for offering suggestions to enhance the readability of our paper. 
\bibliography{ref}

\section*{Checklist}


 \begin{enumerate}

 \item For all models and algorithms presented, check if you include:
 \begin{enumerate}
 \item A clear description of the mathematical setting, assumptions, algorithm, and/or model. [Yes:~See~\Cref{sec:preliminary}, \Cref{sec:method},~\Cref{sec:preliminary_dg}~]
   \item An analysis of the properties and complexity (time, space, sample size) of any algorithm. [Yes:~See~\Cref{sec:related_work}~]
   \item (Optional) Anonymized source code, with specification of all dependencies, including external libraries. [Not Applicable to camera-ready submission]
 \end{enumerate}

 \item For any theoretical claim, check if you include:
 \begin{enumerate}
   \item Statements of the full set of assumptions of all theoretical results. [Yes:~See~\Cref{sec:method}~]
   \item Complete proofs of all theoretical results. [Yes:~See~\Cref{app:proof}~]
   \item Clear explanations of any assumptions. [Yes:~See~\Cref{sec:method}~]     
 \end{enumerate}

 \item For all figures and tables that present empirical results, check if you include:
 \begin{enumerate}
   \item The code, data, and instructions needed to reproduce the main experimental results (either in the supplemental material or as a URL). [Yes:~See~\Cref{sec:exp},~\Cref{subsec:exp_setting}~]
   \item All the training details (e.g., data splits, hyperparameters, how they were chosen). [Yes:~See~\Cref{subsec:exp_setting}~]
   \item A clear definition of the specific measure or statistics and error bars (e.g., with respect to the random seed after running experiments multiple times). [Yes: See caption of~\Cref{fig:benchmark_pacs_diva}~]
   \item A description of the computing infrastructure used. (e.g., type of GPUs, internal cluster, or cloud provider). [Yes:~See~\Cref{app:resources}~]
 \end{enumerate}

 \item If you are using existing assets (e.g., code, data, models) or curating/releasing new assets, check if you include:
 \begin{enumerate}
   \item Citations of the creator If your work uses existing assets. [Yes:~See~\Cref{subsec:exp_setting},~\Cref{app:Licenses}~]
   \item The license information of the assets, if applicable. [Yes:~See~\Cref{app:Licenses}~]
   \item New assets either in the supplemental material or as a URL, if applicable. [Yes:~See~\Cref{sec:exp}~]
   \item Information about consent from data providers/curators. [Not Applicable]
   \item Discussion of sensible content if applicable, e.g., personally identifiable information or offensive content. [Not Applicable]
 \end{enumerate}

 \item If you used crowdsourcing or conducted research with human subjects, check if you include:
 \begin{enumerate}
   \item The full text of instructions given to participants and screenshots. [Not Applicable]
   \item Descriptions of potential participant risks, with links to Institutional Review Board (IRB) approvals if applicable. [Not Applicable]
   \item The estimated hourly wage paid to participants and the total amount spent on participant compensation. [Not Applicable]
 \end{enumerate}
 \end{enumerate}
\section*{Appendix}
\appendix
\section{Mathematical proofs}\label{app:proof}
In this seciton, we restate theoretical results in the main text with the same statement index and provide proofs.
\LabelRestatebleParetoDescentViaConstraintOptShrinkBound*
\begin{proof}\label{proof:constraint_opt_with_bound_shrink_constraint_lead2_Pareto}
Here $s_k>0$ is the number of iterations maintaining constraint in~\Cref{eq:not_feasibility_r}, and $s_k+m_k$ is the number of iterations to ensure the decrease of $\ell$ subject to decreasing $R$ in~\Cref{eq:feasibility_r}.

Note that $b^{(k+j)}$ remains fixed before $s_k$, i.e.~$b^{(k)}=\cdots=b^{(k+s_k+m_k-1)}$ due to~\Cref{eq:b_pareto_descent}. Between $s_k$ till $s_k+m_k$, due to \Cref{eq:b_pareto_descent}, $R$ has decreased.

Due to~\Cref{eq:feasibility_r}, the new reference bound $b^{(k+s_k)}$ (setpoint) has been respected, so $R^{(k+s_k+j)}$ with $0\le j \le m_k$ still remains better than $R^{(k)}$, even though it can be worse than intermediate values $R^{(k+s_k+j^{'})}$ with $0<j^{'}<j$.

Now due to~\Cref{eq:b_pareto_descent}, the minimization of $\ell^{(k+s_k+m_k)}$ cannot lead to values worse than $\ell^{(k)}$ because otherwise~\Cref{eq:feasibility_r} cannot be respected.

Thus both $\ell$ and $R$ decrease at step $k+s_k+m_k$ compared to step $k$.

\end{proof}
\LabelRestatebleParetoDescentViaConstraintOptShrinkBoundCoro*
\begin{proof}\label{proof:coro_constraint_opt_with_bound_shrink_constraint_lead2_Pareto}
  Feasibility in~\Cref{eq:l_decrease} and the condition in~\Cref{eq:feasibility_r} ensures Pareto-descent in~\Cref{def:pareto-descent-operator}, following the proof in~\Cref{prop:pareto_descent_constrained}.
\end{proof}
\LabelRestatebleRegSlide*
\begin{proof}\label{proof:reg_slider}
Multiply $-1$ to both sides of \Cref{eq:deteriorate_old_lambda}, add the results to \Cref{eq:improve_new_lambda}, 
we have
\begin{equation}
(\mu^{(k+1)}-\mu^{(k)})R(\theta^{(k+1)})\le (\mu^{(k+1)}-\mu^{(k)})R({\theta}^{(k)})
\end{equation}
which implies
\begin{equation}
(\mu^{(k+1)}-\mu^{(k)})\left(R(\theta^{(k+1)})-R({\theta}^{(k)})\right)\le 0
\end{equation}
Since $\mu^{(k+1)}>\mu^{(k)}$
\begin{equation}
R(\theta^{(k+1)})-R({\theta}^{(k)}) \le 0
\end{equation}
Equation~\ref{eq:deteriorate_old_lambda} and~\ref{eq:improve_new_lambda} give
\begin{align}
&0\le \mu^{(k)}( R({\theta}^{(k)})  -  R(\theta^{(k+1)})) \le \\
&\ell(\theta^{(k+1)}) -\ell({\theta}^{(k)})  \le  \mu^{(k+1)}( R({\theta}^{(k)})  -  R(\theta^{(k+1)})) 
\end{align}

\end{proof}
\LabelRestatebleEllBoundTelescope*
\begin{proof}\label{proof:reg_ell_bound}
    For $k\in \mathcal{K}_>$ (\Cref{eq:deteriorate_old_lambda} holds), from \Cref{corollary:reg-slide} we have that
    \begin{align*}
        \ell(\theta^{(k+1)})-\ell(\theta^{(k)})&\le \mu^{(k+1)}( R(\theta^{(k)})  -  R(\theta^{(k+1)})).
    \end{align*}
    For $k\in \mathcal{K}_<$, it holds that
    \begin{align*}
        \ell(\theta^{(k+1)}) + \mu^{(k)} R(\theta^{(k+1)}) <  \ell(\theta^{(k)}) + \mu^{(k)} R(\theta^{(k)}),
    \end{align*}
    implying
    \begin{align*}
        \ell(\theta^{(k+1)}) - \ell(\theta^{(k)})   <   \mu^{(k)} (R(\theta^{(k)}) - R(\theta^{(k+1)})).
    \end{align*}
    Decomposing
    \begin{align*}
         &\ell(\theta^{(B)}) - \ell(\theta^{(0)}) \\
         &= \sum_{k = 0}^{B-1} [\ell(\theta^{(k+1)}) - \ell(\theta^{(k)})]\\
         & = \Big(\sum_{k\in\mathcal{K}_>} + \sum_{k\in\mathcal{K}_<}\Big) [\ell(\theta^{(k+1)}) - \ell(\theta^{(k)})]
    \end{align*}
    yields the conclusion.

\end{proof}

\section{Related work and discussion}\label{sec:related_work}
In this section, we discuss the advantages of our method compared to other multiplier scheduling and multi-objective optimization methods, especially in time and space (memory) complexity. Compared to hyperparameter tuning including Bayesian optimization as discussed in~\Cref{remark:pgm_bo_fbopt_realization}, since in general the next multiplier has to be generated after one training round, in the extreme worst case, the number of searches reduces to grid search of $B_d^d$ number of combinations (where~$B_d$ is the number of grid points for each multiplier component, $d$ is the dimension of the multiplier). Our method however requires only a single run. 
\newcommand{\nutanpapersetpoint}{constraint bound}
\subsection{Multiplier adaptation in deep learning}
Multipliers warmup \citep{sonderby2016ladder,ilse2020diva} or phase-in \citep{sicilia2023domain} requires ultimate multiplier values for each loss term, thus necessitates hyperparameter (ultimate multiplier values) optimization, see also analysis in~\Cref{subsec:feedforward_scheduler}.
Other works, such as~\citep{rezende2018taming, klushyn2019learning}, has introduced constraint optimization by representing the weights of loss terms using Lagrange multipliers, designed to prevent over-regularisation of a single term within the loss function, which is extended to multiple loss terms in \citep{chen2022local}. In addition, similar multiplier adaptation schemes were proposed by \citep{shao2020controlvae}, who only dealt with two loss terms, and by \citep{chen2022local}, where the multiplier changes at a rate proportional to the distance of the corresponding loss component from a setpoint, as discussed in \Cref{sec:pid}. However, their methods
relied on a fixed \nutanpapersetpoint~for each loss term and fixed proportionate gain for each loss term (defined to be $K_I$ in \Cref{eq:pid} in our work), necessitating a hyperparameter search, which is challenging when the number of terms in the loss increases. Moreover, for some regularizers, such as the ones based on KL divergence, it is challenging to estimate the \nutanpapersetpoint~range. 
Specifically in the field of domain generalization, it is not obvious how to choose the \nutanpapersetpoint~value and other hyperparameters due to the lack of observations from the target domain, where the validation set from the training domain can easily reach a saturated value
due to the high expressive power of modern neural networks.

In comparison, we deal with many loss terms with multidimensional multipliers and introduced \Cref{eq:setpoint_ada_abs} to progressively adapt the \nutanpapersetpoint~and \Cref{remark:choiceK} to choose the multidimensional controller gain value, thus avoid exhaustive hyperparameter search. 
In addition to the control theory formulation for multiplier adjustment, we also provided a probabilistic graphical model interpretation in \Cref{sec:pgm} and a landscape scheduling interpretation in \Cref{fig:landscape_scheduling}. Furthermore, we discussed our method under the multi-objective optimization scheme compared to single-objective consideration from other works.
Different from the single layer controller in \citep{chen2022local}, we proposed a hierarchical controller in \Cref{sec:pid} to break the multi-objective optimization problem into a series of constraint optimization sub-problems, each configured with a self-adaptive multi-objective setpoint updated via Pareto dominance.

\subsection{Multi-objective optimization for large-scale neural networks}\label{sec:related_multi_obj}
Earlier multi-objective deep learning~\citep{sener2018multi, lin2019pareto} suffers from increased computational cost with the 
number of losses, explicit computation of the gradient with respect to each loss term and gradient norms, making it practically infeasible~\cite{mahapatra2021exact} for large neural networks and large datasets. 
For instance, Pareto Invariance Risk Minimization (PAIR)~\citep{chen2022pareto} studies domain generalization with the use of the multi-objective optimization technique Exact Pareto Optimal Search (EPOS)~\citep{mahapatra2021exact}, which requires explicit full gradient information for each loss term and computation of the $C$ matrix with complexity $m^2n$~\citep{mahapatra2021exact}. With $m$ being the number of loss terms (equivalent to $d+1$ in our paper), and $n$ the dimension of gradient, which is super high in modern neural networks (e.g. 200 million for ResNet50), thus results in heavy memory requirements and excessive computation burden, and hinders the deployment to large-scale networks~\citep{chen2022pareto}.
The Dirichlet sampling of preference vectors was introduced, and the aggregated optimal was computed under the sampled preference vectors \citep{ruchte2021scalable}. However, their sampling of preference vectors introduced data replications. In addition, their method needs to augment the preference vector with the original input, which added complexity of applying their method to different data modes. For instance, for image input, they need first to transform the preference vector to image mode via transpose convolution. Further, their method needs an extra hyperparameter to weight the additional penalty loss to force the solution to obey the preference vectors. 

In comparison, our multi-objective optimization method (M-HOF-Opt) need only $d$ scalar multiplications, thus the extra computation is negligible and can be used in large-scale neural networks. 
EPOS~\cite{mahapatra2021exact} also requires the user to provide a preference vector 
which can be difficult to set, say with 6 objectives for our case. 
In addition,~\citet{chen2022pareto} reported hyperparameter tuning for the preference vector and step length is needed for the method to work properly. 
In comparison, our method, M-HOF-Opt is preference vector free. It ensures the Pareto descent due to the update rule of our setpoint. There is no requirement to search for step length.

Note that the EPOS method~\cite{mahapatra2021exact} is searching for Exact Pareto optimal~\citep{mahapatra2021exact}, which might not exist.
~\citet{chen2022pareto} divides the training into two phases where the first phase uses single-objective optimization favoring ERM loss while the second phase is for balancing different loss terms. However, how many computation resources should one set for the first phase is not clear. Furthermore, the first stage (phase) can affect the exact Pareto optimality.

\section{Experimental details}
\subsection{Experimental Setting}\label{subsec:exp_setting}
This section presents our experiments training the domain generalization model DIVA \citep{ilse2020diva} with $6$ loss terms defined in \Cref{eq:loss_diva_srm} on the widely used domain generalization benchmark dataset PACS \citep{li2017deeper} with different training schemes. Due to the excessive memory requirements and heavy computation of earlier multi-objective deep learning methods discussed in~\Cref{sec:related_work}, which hinders their application, we compare our training scheme with other multiplier scheduling techniques.  
Given that the \textit{sketch} domain within the PACS dataset is considered the most inherently challenging, it was chosen as the sole leave-one-out domain to test the performance of domain generalization. 

To further show the power of~M-HOF-Opt~in automatically balancing different loss terms, we combined two domain generalization algorithms IRM~\citep{arjovsky2019invariant} and DIAL~\citep{levi2021domain}. 

In our experiment, we used the Facebook Research version of ResNet from DomainBed\footnote{\url{https://github.com/facebookresearch/DomainBed/blob/main/domainbed/networks.py}}~\citep{gulrajani2020search} with a learning rate of 5e-5 and batch size of 32.

We first introduce domain generalization and detail the loss formulation of the above mentioned methods and their loss combination, then we present the benchmark setting in~\Cref{sec:benchmark_setting_diva},~\Cref{sec:irm_dial_bench} and the corresponding experimental results. 

\subsection{Domain generalization structral risk}\label{sec:preliminary_dg}
Domain generalization aims at enabling the trained neural network to achieve robust generalization to unseen domains exhibiting distribution shifts~\citep{gulrajani2020search,sun2019variational}. Many domain generalization methods \citep{ganin2016domain,li2018learning,levi2021domain, carlucci2019domain, ilse2020diva, sun2021hierarchical, rame2022fishr} promotes this goal via domain invariant representation by adding domain invariant regularization losses $R(\cdot)$ upon task-specific losses $\ell(\cdot)$ (e.g.~classification loss) on the training data in \Cref{eq:srm}.
\subsubsection{Domain Invariant Variational Autoencoding (DIVA)}\label{subsec:math_diva}
As an example of $\ell(\cdot) + \mu^T R(\cdot)$ loss for domain generalization, we label each component of the loss of Domain Invariant Variational Autoencoding (DIVA)~\citep{ilse2020diva} in the following equations, where we use $x$ to denote the input instance (e.g.~input image), $y$ to denote the corresponding supervision class label, and $d$ the domain label. DIVA~\citep{ilse2020diva} aims to disentangle domain specific information with learned features $z_d$ and class specific information with $z_y$, and has two corresponding classification losses $\mathbb{E}_{q_{\phi_y}(z_y|x)}[\log q_{\omega_y}(y|z_y)]$ (classifying the correct class label based on $z_y$) and $\mathbb{E}_{q_{\phi_d}(z_d|x)}[\log q_{\omega_d}(d|z_d)]$ (classifying the correct domain label based on $z_d$). 

\begin{align}
&L(\theta, \mu, x,y,d)= \ell(\cdot) + \mu^T R(\cdot) \label{eq:loss_diva_srm}\\
&= \gamma_y \mathbb{E}_{q_{\phi_y}(z_y|x)}[\log q_{\omega_y}(y|z_y)]\label{eq:loss_diva_gamma_y} + \\
&\mu_{recon}\mathbb{E}_{q_{\phi_d}(z_d|x), q(z_x|x), q(z_y|x)}\log p_{\theta_r}(x|z_d, z_x, z_y) \label{eq:loss_diva_recon}\\
&\quad + [-\beta_x KL(q_{\phi_x}(z_x|x)||p_{\theta_x}(z_x))] \label{eq:loss_diva_beta_x}\\
&\quad + [-\beta_y KL(q_{\phi_y}(z_y|x)||p_{\theta_y}(z_y|y))] \label{eq:loss_diva_beta_y}\\ 
&\quad + [-\beta_d KL(q_{\phi_d}(z_d|x)||p_{\theta_d}(z_d|d))] \label{eq:loss_diva_beta_d}\\ 
&\quad + \gamma_d \mathbb{E}_{q_{\phi_d}(z_d|x)}[\log q_{\omega_d}(d|z_d)]\label{eq:loss_diva_gamma_d}
\end{align}

Here, $KL$ denotes Kullback-Leibler divergence, $\mathbb{E}$ denotes expectation, $p(\cdot)$ stands for the prior distribution or the distribution of the generative model, and $q(\cdot)$ stands for the approximate posterior distribution. For more details, refer to~\citep{ilse2020diva}.
In the above situation, we have
$\mu=[\mu_{recon}, \beta_x, \beta_y, \beta_d, \gamma_d]$ and $\theta=[\phi_y,\phi_d,\theta_r,\theta_d,\theta_x,\theta_y,\phi_x,\phi_y,\phi_d, \omega_d, \omega_y]$ with each entry representing the weight (with bias) of a neural network. 

\begin{figure*}[ht!]
\centering
\includegraphics[width=0.7\linewidth]{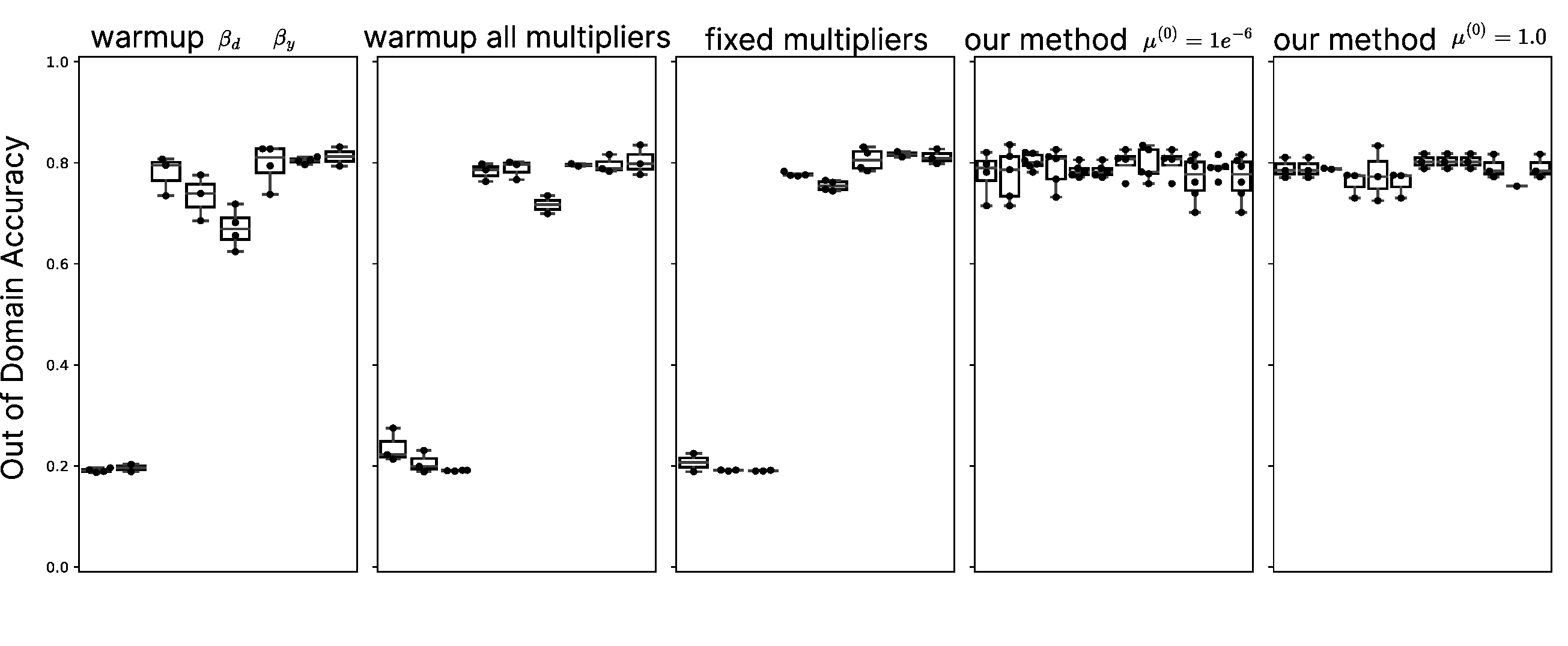}
\caption{Our automatic multiplier adjustment scheme ensures robust out-of-domain test accuracy when training DIVA with the modified \textit{ResNet50} \citep{gulrajani2020search} on PACS dataset (testing on domain \textit{sketch} and training on domain \textit{photo}, \textit{art-painting} and \textit{cartoon}).
Each panel (i.e., subplot) corresponds to a training scheme. 
Each box plot inside a panel corresponds to a specific hyperparameter combination (i.e.~multipliers value for baselines and controller hyperparameters for M-HOF-Opt).
Inside each box plot, we repeat the experiment with different random seeds corresponding to the dots scatter.
   In the first panel, we warm up only $\beta_y$, $\beta_d$ in \Cref{eq:loss_diva_beta_d,eq:loss_diva_beta_y} to their ultimate value $1.0$ while sampling fixed multiplier $\gamma_d$ and $\gamma_y$ while fixing $\mu_{recon}=1.0$. 
In the second panel,  additionally, we also warm up $\mu_{recon}$ in \Cref{eq:loss_diva_recon} to $1.0$ and warm up $\gamma_d$ in \Cref{eq:loss_diva_gamma_d} to the sampled value, and keep $\gamma_y$ fixed to be the sampled value.
In the third panel, we use fixed constant multipliers without warm-up while keep $\beta=1.0$ and $\mu_{recon}=1.0$ and only sample $\gamma_d,\gamma_y$ combinations. The last two panels correspond to results utilizing our multi-objective hierarchical feedback optimization training method, and we let $\mu$ include all multipliers except $\gamma_y=1.0$ in \Cref{eq:loss_diva_gamma_y}. For simplicity, when we write $\mu^{(0)}=1.0$ where $\mu^{(0)} \in \mathbb{R}_{+}^d$, we mean each component of $\mu^{(0)}$ equals $1.0$. For reproducibility, see~\url{https://github.com/marrlab/DomainLab/tree/mhof}.
}\label{fig:benchmark_pacs_diva}
\end{figure*}

In~\citep{ilse2020diva}, $\mu_{recon}=1.0$ in~\Cref{eq:loss_diva_recon}. Additionally, $\gamma_y$ and $\gamma_d$ are maintained as constants. $\gamma_y$ is corresponding to $\ell(\cdot)=\gamma_y\mathbb{E}_{q_{\phi_y}(z_y|x)}[\log q_{\omega_y}(y|z_y)]$ in~\Cref{eq:loss_diva_gamma_y}, and $\gamma_d$ in~\Cref{eq:loss_diva_gamma_d} is associated with one component of $R(\cdot)$. However, the combined choice of $\gamma_y$ and $\gamma_d$ significantly influences the generalization performance, as shown by our experimental findings in~\Cref{fig:benchmark_pacs_diva}. 

\begin{remark}[feedforward scheduler]\label{subsec:feedforward_scheduler}
For the multiplier $\beta_x, \beta_y, \beta_d$ corresponding to other components of $R(\cdot)$,
\citet{ilse2020diva} used a warm-up strategy to increase them gradually from a small value to a pre-defined value. This, however, still requires a choice of the ultimate values for $\beta_x, \beta_y, \beta_d$. In~\citep{ilse2020diva}, these ultimate values are simply set to one. We coin this kind of multiplier scheduling strategy a \emph{feedforward} scheme. Note that this strategy does not reduce the number of hyperparameters to be selected, since ultimate values for the multipliers still have to be specified. 
\end{remark}

\subsubsection{Invariant Risk Minimization (IRM) and Domain Invariant Adversarial Learning (DIAL)}\label{subsec:math_irm_dial}
Let $\Phi$ be the feature extraction neural network, $w$ be the task network (e.g.~classifier), let $d$ index available training domains. Invariant Risk Minimization (IRM) uses the following optimization:
\begin{align}
&\min_{\Phi, w} \sum_{d} \ell^{(d)}(w \circ \Phi(X)) + \nonumber\\
&\mu \sum_{d} \|\nabla_{w|w=1.0} \ell^{(d)}(w \circ \Phi(X))\|^2   
\end{align}
while Domain Invariant Adversarial Learning proposed using adversarial perturbed image ($X_{adv}$ in comparison to original data $X$) as new training samples to form adversarial loss. If we combine these two, 
we essentially get the following loss:
\begin{align}
&\ell + \mu^T R \nonumber\\
=&\min_{\Phi, w} \sum_{d} \ell^{(d)}(w \circ \Phi(X)) + \nonumber\\
&\mu_1 \sum_{d} \|\nabla_{w|w=1.0} \ell^{(d)}(w \circ \Phi(X))\|^2 + \nonumber\\ &\mu_2\ell(w\circ \Phi (X_{adv})) \label{eq:irm_dial} 
\end{align}

\subsection{Benchmark settings for DIVA}\label{sec:benchmark_setting_diva}
In the benchmark, we sample different combinations of hyperparameters, (including multiplier for baselines and controller hyperparameter for our method) for each method to be compared. 
For baselines, we sample $\gamma_d\in \{1, 1001, 100001\}$ and $\gamma_y\in \{1, 1001, 100001\}$ value combinations, and set the ultimate values of $\beta_y,\beta_d$ to be $1.0$, following \citep{ilse2020diva}.  Our method does not need to set ultimate values for $\mu$, so we sample controller hyperparameters, 
as well as different initial conditions $\mu^{(0)}$ (Each component of the $\mu^{(0)}$ vector set to be the same value). 

\subsection{Benchmark settings and results for IRM and DIAL}\label{sec:irm_dial_bench}
\begin{figure}[htbp] 
\centering
\includegraphics[width=0.8\linewidth]{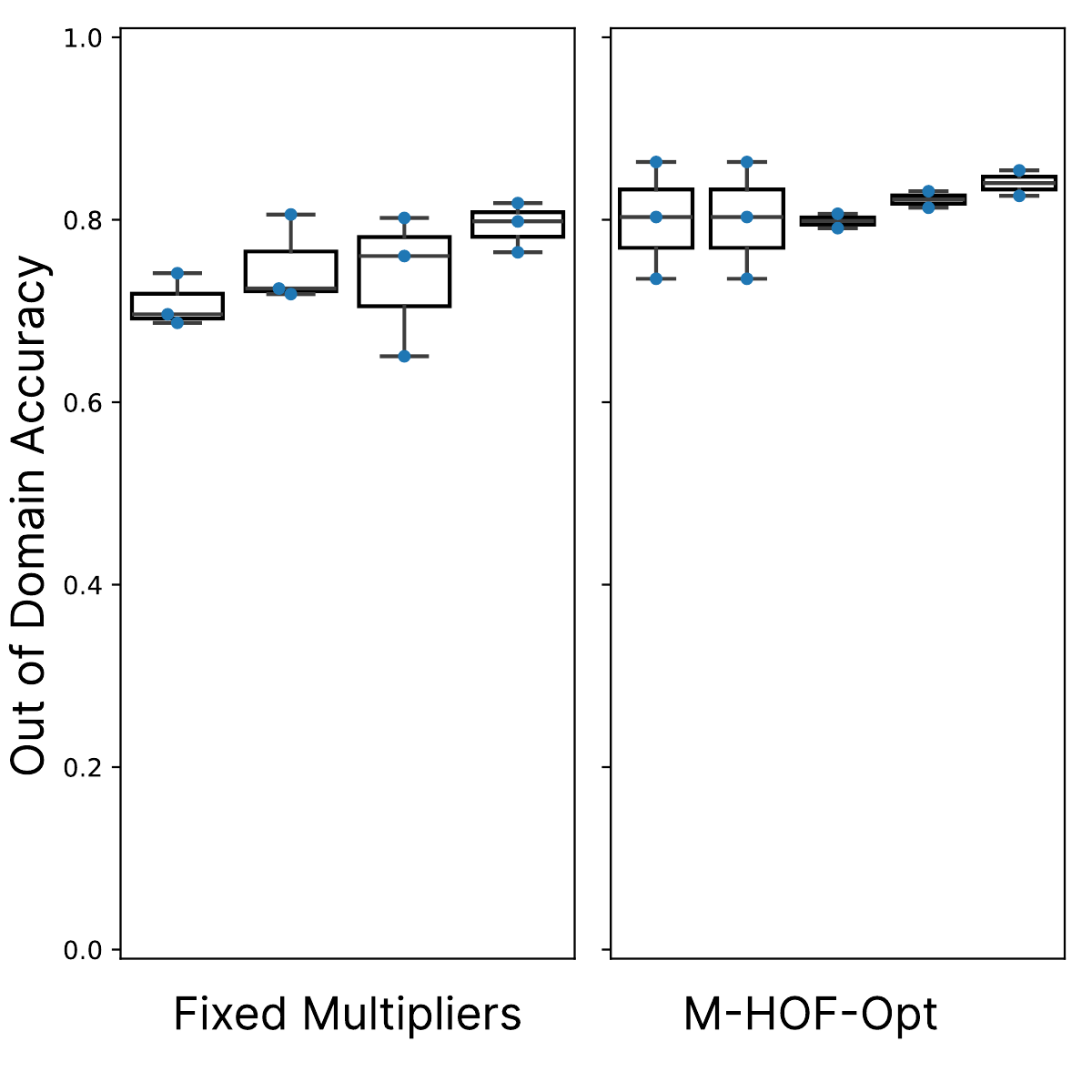}\caption{M-HOF-Opt achieves better out-of-domain generalization accuracy compared to using fixed multipliers, when combining  regularization loss from Domain Invariant Adversarial Learning (DIAL) and Invariant Risk Minimization (IRM) in~\Cref{subsec:math_irm_dial}. Each panel (i.e., subplot) corresponds to a training scheme. 
Each box plot inside a panel corresponds to a specific hyperparameter combination (i.e.~multiplier values for baseline ``Fixed Multipliers'' and controller hyperparameters for M-HOF-Opt). Inside each box plot, we repeat the experiment with different random seeds corresponding to the dots scatter. From the figure, we can see that the baseline performance varies a lot as the multiplier combination changes while M-HOF-Opt remains robust against controller hyperparameter variations. See~\Cref{subsec:exp_setting,sec:irm_dial_bench} for benchmark setting. 
}\label{fig:bench_irm_dial}
\end{figure}

For baseline (named ``Fixed Multipliers'' in~\Cref{fig:bench_irm_dial}), we choose $\mu_1,\mu_2$ in~\Cref{eq:irm_dial} from $0.01, 0.1, 1.0, 10$.

For our method M-HOF-Opt, we choose different controller hyperparameters, with $\mu^{(0)}=1e^{-6}, 0.001$, $\eta=0.325, 0.775$ (uniform between 0 and 1), $\mu^{(clip)}=10,100,1000$.

In~\Cref{fig:bench_irm_dial}, each box plot inside the rectangle correspond to a hyperparameter configuration. For baseline ``Fixed Multipliers'' in~\Cref{fig:bench_irm_dial}, this corresponds to $\mu_1, \mu_2$ values in~\Cref{eq:irm_dial}.
For M-HOF-Opt, each box plot corresponds to different controller hyperparameters. 

From~\Cref{fig:bench_irm_dial}, we can see that M-HOF-Opt 
performs robust on out-of-domain generalization against controller hyperparameter changes, while when searching for fixed $\mu_1, \mu_2$ combinations in~\Cref{eq:irm_dial} lead to worse results. 


\subsection{Effects of different low level optimizers}\label{sec:adamw_cosineLR}
In this section, we investigate if our algorithm M-HOF-Opt still works when using different low level (gradient descent) optimization algorithms. Using the same setting as in~\Cref{sec:irm_dial_bench}, we obtain~\Cref{tab:methods_comparison_adam_w} for changing \textit{Adam} with \textit{AdamW}~\citep{loshchilov2017fixing} and~\Cref{tab:methods_comparison_lr_anneal} for stacking on top of the low level optimization algorithm with \textit{CosineAnnealingLR}\footnote{\url{https://pytorch.org/docs/stable/generated/torch.optim.lr_scheduler.CosineAnnealingLR.html}}. The results show that M-HOF-Opt is robust against changes of low level optimization algorithms.
\begin{table}[h!]
\centering
\begin{tabular}{|l|c|c|}
\hline
\textbf{Method} & \textbf{Mean} & \textbf{Std} \\ \hline
Warmup & 0.767 & 0.0467 \\ \hline
Fixed Multipliers & 0.748 & 0.052 \\ \hline
M-HOF-Opt & 0.806 & 0.0456 \\ \hline
\end{tabular}
\caption{M-HOF-Opt still achieves better out-of-domain generalization accuracy when replacing Adam with AdamW as low level optimizer. For benchmark and experimental setting, see~\Cref{subsec:exp_setting,sec:adamw_cosineLR}.}\label{tab:methods_comparison_adam_w}
\end{table}

\begin{table}[h!]
\centering
\begin{tabular}{|l|c|c|c|}
\hline
\textbf{Method} & \textbf{Mean} & \textbf{Std} \\ \hline
Warmup & 0.750 & 0.0441\\ \hline
Fixed Multipliers & 0.7322 & 0.0570\\ \hline
M-HOF-Opt & 0.7873 & 0.0326 \\ \hline
\end{tabular}
\caption{M-HOF-Opt still achieves favorable out-of-domain generalization accuracy when stacking \textit{CosineAnnealingLR} as another hierarchy on top of the low level gradient based optimization algorithm. M-HOF-Opt treats the stacked \textit{CosineAnnealingLR} and lower level gradient optimization algorithm as the uncontrolled plant. See~\Cref{subsec:exp_setting,sec:adamw_cosineLR} for benchmark and experimental setting.}\label{tab:methods_comparison_lr_anneal}
\end{table}

\section{Computer resources}\label{app:resources}
All experiments were conducted on our internal computation clusters with NVIDIA V100 GPUs.

\section{Licenses for Existing Assets}\label{app:Licenses}

\paragraph{Code}
Our implementation leverages several open-source libraries:
\begin{itemize}
    \item \textbf{PyTorch}: BSD 3-Clause License. \url{https://pytorch.org/}
    \item \textbf{NumPy}: BSD 3-Clause License. \url{https://numpy.org/}
    \item \textbf{Matplotlib}: Matplotlib License.
    \url{https://matplotlib.org/}
\end{itemize}
In our experiment, we used the Facebook Research version of ResNet from DomainBed (\url{https://github.com/facebookresearch/DomainBed/blob/main/domainbed/networks.py}) which is MIT License.

\paragraph{Data}
We used the PACS dataset for non-commercial research purposes and we cited the author. 

\section{Contributions} 
XS proposed the idea, developed the theories and algorithms, implemented the code, designed and carried out the experiments, processed experimental data and generated the figures, wrote the manuscript. NC proposed the PI-like multiplier adaptation in \Cref{sec:pid}, helped XS with PI-like controller implementation, held various discussions, proofread and improved the manuscript.  AG initiated the visualization code for \Cref{fig:dyn_pacs_diva_fbopt} and \Cref{fig:phase_portrait_pacs_diva_fbopt}, refined by XS, proofread and improved the manuscript. YX developed \Cref{corollary:ellbound} and corresponding remarks with XS, proofread and improved the  constraint optimization part of the theory, discussed on control theory part. CF initiated the benchmark code with refinements from XS, proofread \Cref{corollary:reg-slide}. MW improved the code for \textit{DomainLab}, especially M-HOF-Opt. ED and FD tested and improved the code for benchmark, proofread the manuscript. FD further improved the \textit{tensorboard} visualization of the algorithm. LB proofread \Cref{corollary:reg-slide}, tested the benchmark code. DS tested the benchmark code, assisted XS in collecting experimental results. CM supervised the project, offered extensive proofreads of the manuscript. 

\end{document}